\documentclass[10pt,twocolumn,letterpaper]{article}

\usepackage[pagenumbers]{cvpr} 

\usepackage{graphicx}
\usepackage{amsmath}
\usepackage{amssymb}
\usepackage{booktabs}
\usepackage{mathtools}
\usepackage{multirow}
\usepackage{enumitem}
\usepackage{dsfont}
\usepackage[accsupp]{axessibility}  

\DeclareMathOperator*{\argmin}{arg\,min}

\usepackage[pagebackref,breaklinks,colorlinks]{hyperref}

\usepackage[capitalize]{cleveref}
\crefname{section}{Sec.}{Secs.}
\Crefname{section}{Section}{Sections}
\Crefname{table}{Table}{Tables}
\crefname{table}{Tab.}{Tabs.}
\Crefname{algorithm}{Algorithm}{Algorithms}
\crefname{algorithm}{Alg.}{Algs.}

\usepackage{algorithm}
\usepackage[noend]{algpseudocode}

\usepackage{afterpage}

\def\cvprPaperID{2050} 
\def\confName{CVPR}
\def\confYear{2023}

\begin{document}

\title{Regularizing Second-Order Influences for Continual Learning}

\author{Zhicheng Sun$^1$, Yadong Mu$^{1,2}$\thanks{Corresponding author.} , Gang Hua$^3$\\
$^1$Peking University, $^2$Peng Cheng Laboratory, $^3$Wormpex AI Research\\
{\tt\small \{sunzc,myd\}@pku.edu.cn, ganghua@gmail.com}
}
\maketitle

\begin{abstract}
Continual learning aims to learn on non-stationary data streams without catastrophically forgetting previous knowledge. Prevalent replay-based methods address this challenge by rehearsing on a small buffer holding the seen data, for which a delicate sample selection strategy is required. However, existing selection schemes typically seek only to maximize the utility of the ongoing selection, overlooking the interference between successive rounds of selection. Motivated by this, we dissect the interaction of sequential selection steps within a framework built on influence functions. We manage to identify a new class of second-order influences that will gradually amplify incidental bias in the replay buffer and compromise the selection process. To regularize the second-order effects, a novel selection objective\linebreak is proposed, which also has clear connections to two widely adopted criteria. Furthermore, we present an efficient implementation for optimizing the proposed criterion. Experiments on multiple continual learning benchmarks demonstrate the advantage of our approach over state-of-the-art methods. Code is available at \url{https://github.com/feifeiobama/InfluenceCL}.
\end{abstract}

\begin{figure}
    \centering
    \begin{subfigure}{\linewidth}
        \refstepcounter{subfigure}\label{fig:selection_a}
        \refstepcounter{subfigure}\label{fig:selection_b}
    \end{subfigure}
    \includegraphics[width=\linewidth]{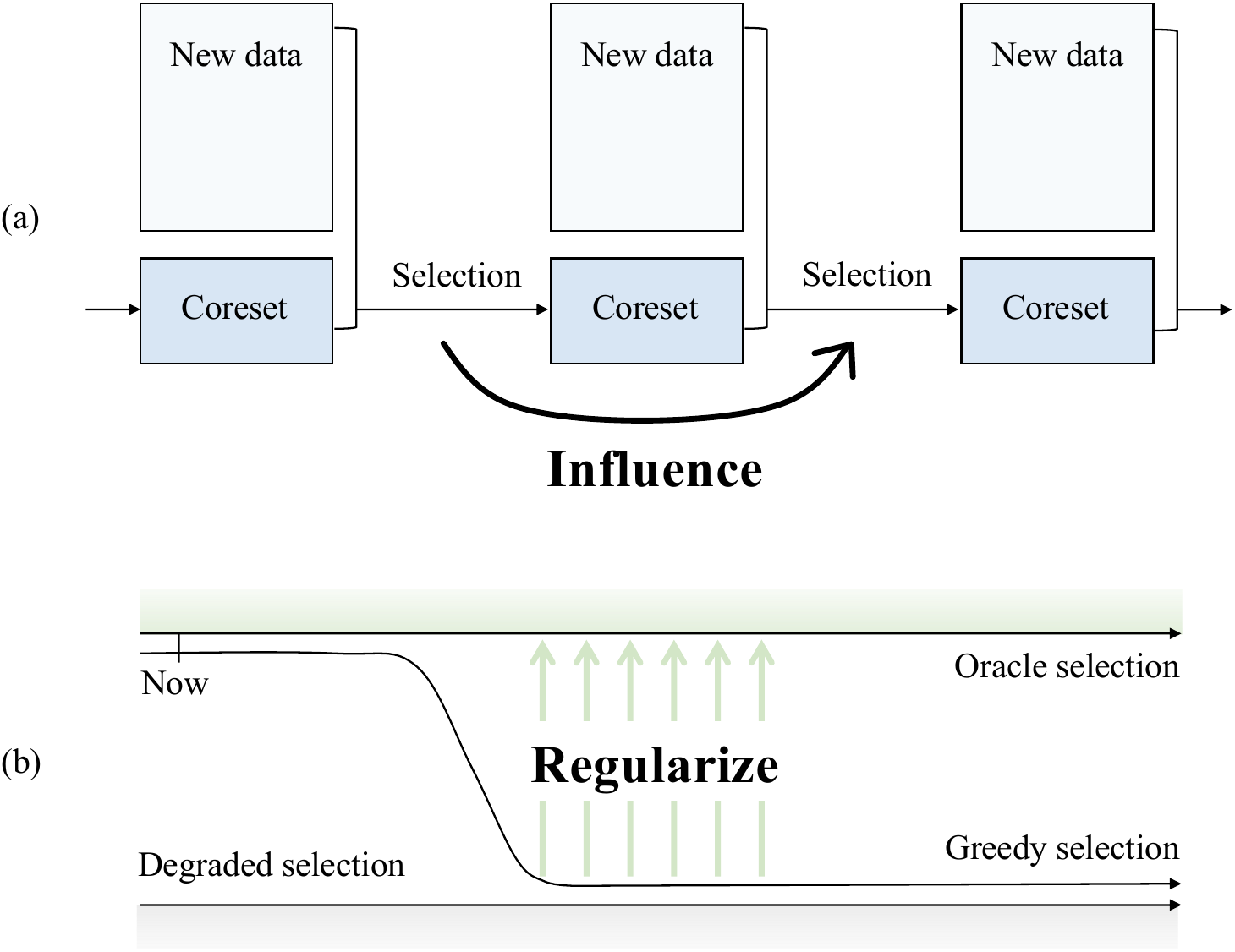}
    \caption{(a) In continual learning, earlier coreset selection exerts a profound influence on subsequent steps through the data flow. (b)~By ignoring this, a greedy selection strategy can degrade over time. Therefore, we propose to model and regularize the influence of each selection on the future.}
    \label{fig:selection}
\end{figure}

\section{Introduction}

The ability to continually accumulate knowledge is a hallmark of intelligence, yet prevailing machine learning systems struggle to remember old concepts after acquiring new ones. \textit{Continual learning} has hence emerged to tackle this issue, also known as \textit{catastrophic forgetting}, and thereby enable the learning on long task sequences which are frequently encountered in real-world applications~\cite{chen2018lifelong, de2021continual}. Amongst a variety of valid methods, the replay-based approaches~\cite{rebuffi2017icarl, lopez2017gradient} achieve evidently strong results by buffering a small \textit{coreset} of previously seen data for rehearsal in later stages. Due to the rigorous constraints on memory overhead, the replay buffer needs to be carefully maintained such that only the samples deemed most critical for preserving overall performance are selected during the learning procedure.

Despite much effort in sophisticated design of coreset selection criteria~\cite{aljundi2019gradient, yoon2022online, tiwari2022gcr, borsos2020coresets}, existing works are mostly developed for an oversimplified scenario, where each round of selection is considered isolated and primarily focused on refining single-round performance. For example, Borsos \etal~\cite{borsos2020coresets} greedily minimize the empirical risk on the currently available subset of data. However, as shown in~\cref{fig:selection_a}, the actual selection process within continual learning is rather different in that each prior selection defines the input for subsequent selection and therefore has an inevitable impact on the later decision. Neglecting such interactions between successive selection steps, as in previous practice, may result in a degraded coreset after the prolonged selection process. To maximize overall performance, an ideal continual learning method should take into account the future influence of each round of selection, which remains unresolved due to the obscure role of intermediate selection results.

This work models the interplay among multiple selection steps with the classic tool of influence functions~\cite{hampel1974influence, koh2017understanding}, which estimates the effect of each training point by perturbing sample weights. Similarly, we begin by upweighting two samples from consecutive time steps, and uncover that the latter one's influence evaluation is biased due to the consequent perturbations in test gradient and coreset Hessian. This gives rise to a new class of second-order influences that interfere with the initial influence-based selection strategy. To be specific, the selection tends to align across different rounds, favoring those samples with a larger gradient projection along the previously upweighted ones. It further suggests that some unintended adverse effects of the early selection steps, such as incidental bias introduced into the replay buffer, will be amplified after rounds of selection and impair the final performance.

To address the newly discovered disruptive effects, we propose to regularize each round of selection beforehand based on its expected second-order influence, as illustrated in~\cref{fig:selection_b}. Intuitively, the selection with a large magnitude of second-order influence will substantially interfere with future influence estimates, and therefore should be avoided. However, the magnitude itself cannot be precalculated for direct guidance of the selection, since it is related to some unknown future data. We instead derive a tractable upper bound for the magnitude which results in the form of $\ell_2$-norm, and integrated it with the vanilla influence functions to serve as the final selection criterion. The proposed regularizer can be interpreted with clarity as it equates to a couple of existing criteria ranging from gradient matching to diversity, under varying simplifications. Finally, we present an efficient implementation that tackles the technical challenges in selecting with neural networks.

Our contributions are summarized as below:
\begin{itemize}
    \itemsep0em 
    \item We investigate the previously-neglected interplay between consecutive selection steps within an influence-based selection framework. A new type of second-order influences is identified, and further analysis states its harmfulness in continual learning.
    \item A novel regularizer is proposed to mitigate the second-order interference. The regularizer is associated with two other popular selection criteria and can be efficiently optimized with our implementation.
    \item Comprehensive experiments on three continual learning benchmarks (Split CIFAR-10, Split CIFAR-100 and Split \textit{mini}ImageNet) clearly illustrates that our method achieves new state-of-the-art performance.
\end{itemize}

\section{Related works}

\textbf{Continual learning} studies the training of models on a sequence of tasks with potential data distribution shift. It is known for suffering from catastrophic forgetting~\cite{mccloskey1989catastrophic}, where the model abruptly forgets past knowledge after being updated on new tasks. To overcome this effect, three main streams of methods have been developed: weight regularization~\cite{kirkpatrick2017overcoming, li2017learning, zeng2019continual}, parameter isolation~\cite{rusu2016progressive, serra2018overcoming, aljundi2017expert, yoon2018lifelong} and memory replay~\cite{rebuffi2017icarl, lopez2017gradient, shin2017continual}. However, the first two approaches exhibit over-regularization or unconstrained parameter growth on long task sequences~\cite{hadsell2020embracing}. In this work, we focus on the replay-based approach, which stores a small subset of previous data in a memory buffer, to be revisited when learning new tasks.

One of the key components in replay-based methods is the sample selection strategy, which undertakes the task of summarizing representative data into the replay buffer. The pioneering works~\cite{lopez2017gradient, rebuffi2017icarl} featured a ring buffer for exemplar management, followed by reservoir sampling~\cite{vitter1985random, riemer2019learning} and its class-balanced variants~\cite{chaudhry2019tiny, prabhu2020gdumb}. More sophisticated selection criteria have been proposed as well, such as matching the overall gradient~\cite{balles2021gradient, tiwari2022gcr} or enhancing the diversity of samples~\cite{aljundi2019gradient, yoon2022online, bang2021rainbow}. Recently, Borsos \etal~\cite{borsos2020coresets} and Sun \etal~\cite{sun2022exploring} introduced influence functions~\cite{koh2017understanding} to continual learning, providing a fresh and interpretable perspective on sample selection. Our method builds on an influence-based framework but also has clear connections to gradient matching and diversity.

\textbf{Influence functions}, known from robust statistics~\cite{hampel1974influence}, were recently advocated in machine learning by Koh and Liang~\cite{koh2017understanding} to estimate the effect of upweighting a training sample to the model parameter and test loss. Since then, influence functions have served as sample selection criteria for various scenarios such as data-efficient learning~\cite{ting2018optimal, wang2020less}, class imbalanced learning~\cite{park2021influence} and noisy label learning~\cite{kong2022resolving}. Typically, samples with the highest influence scores are identified and then downweighted or discarded. Closely related to our work are their applications in continual learning~\cite{borsos2020coresets, zhou2021overcoming, sun2022exploring} and domain adaptation~\cite{chen2020multi}, but they were still developed for the single-round selection scenario described earlier and overlooked the high-order interactions between multiple rounds of selection, leading to suboptimal results in the continual learning setup.

There are also efforts made toward more accurate and efficient calculation of influence functions. For example, Basu~\etal~\cite{basu2020second} introduced second-order approximations to measure group effects~\cite{koh2019accuracy, basu2020second} of large, coherent groups of training points. However, their analysis is limited to jointly optimized samples and cannot be applied to continuously incoming data. In the other direction, various speedup tricks~\cite{guo2021fastif, schioppa2022scaling, borsos2020coresets, zhang2022rethinking} have been proposed for over-parameterized neural networks, among which a representative approach is to use neural tangent kernels~\cite{jacot2018neural} as the proxy model~\cite{borsos2020coresets, zhang2022rethinking}. We adapt some of these techniques to the proposed second-order influence functions and derive an efficient selection criterion for continual learning.

\section{Influence functions for continual learning}
\subsection{Problem formulation}
We consider learning on a continuous stream of data $\mathcal{Z}=\{z_i\}_{i=1}^n$ assuming that only a fraction of samples $\mathcal{Z}_t\subset \mathcal{Z}$ can be accessed at each time step $t$. The main challenge in such a learning paradigm is to retain performance on the seen data $\mathcal{Z}_{1:t}=\bigcup_{i=1}^{t}\mathcal{Z}_i$ in later stages. To achieve this, we adopt a most straightforward and effective approach that stores a few samples $\mathcal{C}_t$ in a replay buffer as the representatives of previous data $\mathcal{Z}_{1:t}$, where the memory size $|\mathcal{C}_t|\le m$ and $m\ll n$. In consequence, it is crucial to maintain a high-quality subset throughout the training process~\cite{yoon2022online, aljundi2019gradient}. Formally, let $\theta$ be the model parameters and $L(z_i,\theta)$ be the loss on training point $z_i$. Then, our selection goal is to preserve model performance on $\mathcal{Z}_{1:t}$ by replaying on $\mathcal{C}_t$, which is formulated as:
\begin{equation}
\label{eq:problem}
\begin{aligned}
\smashoperator[r]{\min_{\mathcal{C}_t\subset\mathcal{C}_{t-1}\cup\mathcal{Z}_t,|\mathcal{C}_t|\le m }}\quad&\qquad\;\smashoperator[r]{\sum_{z_i\in\mathcal{Z}_{1:t}}}\;\;L(z_i,\hat\theta)\\
\textrm{s.t.}\quad&\hat\theta=\argmin_{\theta}\sum_{z_i\in\mathcal{C}_t}L(z_i,\theta).
\end{aligned}
\end{equation}
Though its main form is consistent with that of the well-studied coreset selection problem~\cite{borsos2020coresets}, note that the outer objective is intractable since $\mathcal{Z}_{1:t}$ is partly unavailable.

In the following sections, we first present a baseline selection strategy produced by the vanilla influence functions, then showcase its limitation in handling continual selection and propose our improved version.

\subsection{Influence-based selection}
\label{sect:ifs}
Influence functions~\cite{koh2017understanding} provide an efficient approximation for solving the coreset selection problem by perturbing sample weights, for which the previous exact solution requires expensive leave-one-out retraining.

Before diving in, we first convert~\cref{eq:problem} into a tractable problem by leveraging the empirical risk on $\mathcal{C}_{t-1}\cup\mathcal{Z}_t$\footnote{The loss on $\mathcal{Z}_t$ needs to be reweighted by a constant to balance with the loss on $\mathcal{C}_{t-1}$. We omit the weight for simplicity unless otherwise stated.} as a proxy for the original test loss on $\mathcal{Z}_{1:t}$, presuming a close correlation between them:
\begin{equation}
\label{eq:problem_new}
\begin{aligned}
\smashoperator[r]{\min_{\mathcal{C}_t\subset\mathcal{C}_{t-1}\cup\mathcal{Z}_t,|\mathcal{C}_t|\le m }}\quad&\quad\;\;\smashoperator[r]{\sum_{z_i\in\mathcal{C}_{t-1}\cup\mathcal{Z}_t}}\;\;L(z_i,\hat\theta)\\
\textrm{s.t.}\quad&\hat\theta=\argmin_{\theta}\sum_{z_i\in\mathcal{C}_t}L(z_i,\theta).
\end{aligned}
\end{equation}
Solving this involves uncovering the effect of selecting or discarding some training sample $z\in\mathcal{C}_{t-1}\cup\mathcal{Z}_t$, for which a small weight $\epsilon$ is added to the interested sample, and then the optimal point in the inner optimization becomes:
\begin{equation}
\label{eq:if_first}
\hat\theta_{\epsilon,z}=\argmin_{\theta}\sum_{z_i\in\mathcal{C}_t}L(z_i,\theta)+\epsilon L(z,\theta).
\end{equation}
Let $\hat\theta_t=\hat\theta_{\epsilon,z}|_{\epsilon=0}$ denote the initial optimal parameters and $H_{\hat\theta_t}=\sum_{z_i\in \mathcal{C}_t}\nabla_\theta^2L(z_i,\hat\theta_t)$ denote the Hessian which by assumption is positive definite. A classic result by Cook and Weisberg~\cite{cook1982residuals} yields the change in model parameters as $\frac{d\hat\theta_{\epsilon,z}}{d\epsilon}\Big|_{\epsilon=0}=-H_{\hat\theta_t}^{-1}\nabla_\theta L(z,\hat\theta_t)$, from which we can derive the influence of upweighting $z$ on the outer loss in~\cref{eq:problem_new}:
\begin{equation}
\label{eq:if_loss}
\begin{aligned}
\mathcal{I}(z)&=\smashoperator[r]{\sum_{z_i\in\mathcal{C}_{t-1}\cup\mathcal{Z}_t}}\;\;\frac{dL(z_i,\hat\theta_{\epsilon,z})}{d\epsilon}\Big|_{\epsilon=0}\\
&=-\;\;\smashoperator[lr]{\sum_{z_i\in\mathcal{C}_{t-1}\cup\mathcal{Z}_t}}\;\;\nabla_\theta L(z_i,\hat\theta_t)^\top H_{\hat\theta_t}^{-1}\nabla_\theta L(z,\hat\theta_t).
\end{aligned}
\end{equation}
We denote the inverse Hessian-vector product therein as $s_t=H_{\hat\theta_t}^{-1}\sum_{z_i\in\mathcal{C}_{t-1}\cup\mathcal{Z}_t}\nabla_\theta L(z_i,\hat\theta_t)$ for future use.

To minimize the outer loss, the coreset $\mathcal{C}_t$ should contain the samples with the lowest negative influence scores. It can be efficiently solved by a greedy strategy, which starts from the full set $\mathcal{C}_t=\mathcal{C}_{t-1}\cup\mathcal{Z}_t$ and discards the sample $z$ with the largest influence at each iteration until the memory constraint $|\mathcal{C}_t|\le m$ is met. While such a strategy is qualified in a variety of evaluations~\cite{wang2018data, wang2020less, borsos2020coresets, zhou2021overcoming}, we argue that it is not the answer to continual learning by dissecting the joint effect of influence-based selection in a sequence.

\subsection{Second-order influences in continual selection}
\label{sect:ifs_second}
In continual learning, each prior selection determines the input for subsequent steps and thus influences the effectiveness of future selection, \ie, it may interfere with the later evaluation of the selection criterion.

To model such an interaction, we consider two samples $z$ and $z'$ from consecutive steps $t$ and $t+1$. With the previous sample $z$ upweighted by $\epsilon$, the next round of selection is also affected, by a drift in the influence score of $z'$. Specifically, there are two cases depending on whether $z$ and $z'$ are jointly optimized in the following step:

(1) The samples $z$ and $z'$ are not jointly optimized, meaning that the previous sample $z$ is excluded from the coreset in the next round and only serves as a test point. As a result, only the outer loss in~\cref{eq:problem_new} is affected, with the influence score of the subsequent candidate $z'$ changed to:
\begin{equation}
\resizebox{\linewidth}{!}{$
\begin{aligned}
\mathcal{I}_{\epsilon,z}(z')=-&\Biggl(\smashoperator[r]{\sum_{z_i\in\mathcal{C}_t\cup\mathcal{Z}_{t+1}}}\;\nabla_\theta L(z_i,\hat\theta_{t+1})+\epsilon\nabla_\theta L(z,\hat\theta_{t+1})\Biggr)^\top\\
&H_{\hat\theta_{t+1}}^{-1}\nabla_\theta L(z',\hat\theta_{t+1}).
\end{aligned}
$}
\end{equation}
By taking the derivative \wrt $\epsilon$, it yields the influence of upweighting $z$ on the influence score of $z'$. Alternatively, it can be viewed as the second-order influence of two samples from consecutive selection steps on the outer loss:
\begin{equation}
\label{eq:if_second_1}
\begin{aligned}
\mathcal{I}^{(2)}(z,z')&=\frac{d \mathcal{I}_{\epsilon,z}(z')}{d\epsilon}\Big|_{\epsilon=0}\\
&=-\nabla_\theta L(z,\hat\theta_{t+1})^\top H_{\hat\theta_{t+1}}^{-1}\nabla_\theta L(z',\hat\theta_{t+1}).
\end{aligned}
\end{equation}
We will soon analyze its effect using the geometric view of projection.

(2) The samples $z$ and $z'$ are jointly optimized by the inner objective in~\cref{eq:problem_new}. In this case, the inner Hessian is also changed, in the direction of $H_{\hat\theta_{t+1},z}\coloneqq\nabla_\theta^2L(z,\hat\theta_{t+1})$. Hence, the influence score of the following sample $z'$ is modified to:
\begin{equation}
\resizebox{\linewidth}{!}{$
\begin{aligned}
\mathcal{I}_{\epsilon,z}(z')=-&\Biggl(\smashoperator[r]{\sum_{z_i\in\mathcal{C}_t\cup\mathcal{Z}_{t+1}}}\;\nabla_\theta L(z_i,\hat\theta_{t+1})+\epsilon\nabla_\theta L(z,\hat\theta_{t+1})\Biggr)^\top\\
&(H_{\hat\theta_{t+1}}+\epsilon H_{\hat\theta_{t+1},z})^{-1}\nabla_\theta L(z',\hat\theta_{t+1}).
\end{aligned}
$}
\end{equation}
Similarly, the second-order influence of $z'$ and $z$ can be obtained as follows (see Supplementary for the derivation):
\begin{equation}
\label{eq:if_second_2}
\resizebox{\linewidth}{!}{$
\begin{aligned}
&\mathcal{I}^{(2)}(z,z')=\frac{d \mathcal{I}_{\epsilon,z}(z')}{d\epsilon}\Big|_{\epsilon=0}\\
&=-(\nabla_\theta L(z,\hat\theta_{t+1})-H_{\hat\theta_{t+1},z}s_{t+1})^\top H_{\hat\theta_{t+1}}^{-1}\nabla_\theta L(z',\hat\theta_{t+1}).
\end{aligned}
$}
\end{equation}
Compared to the first case, there is an additional term related to the Hessian perturbation, whose effect will be discussed in~\cref{sect:connection}. Nevertheless, the prior term in~\cref{eq:if_second_1} is fully retained.

\paragraph{Second-order influences compromise sample diversity.} For an intuitive understanding of the new influences, let $\langle u,v\rangle=u^T H_{\hat\theta_{t+1}}^{-1}v$ define an inner product over the gradient space and induce the notion of projection. In both cases, the sample $z'$ with a larger gradient projection onto the previous $z$ is estimated with a lower influence and thus more likely to be selected. This drives the influence-based selection strategy to include more similar samples to the buffer and expel dissimilar ones, leading to a lack of diversity.

\paragraph{Second-order influences amplify memory bias.}
While the above analysis is based on a real perturbation to sample weights, it could also be a simulated perturbation on the gradient statistics of the coreset. Such perturbations are common in the learning process, as each selection step may introduce an incidental bias to the buffer. Like real samples, they can produce the same second-order effect of attracting similar ones into the buffer. Although these perturbations are expected to have zero mean like white noises, the earlier perturbations have taken a greater effect in accumulating memory bias, which results in inaccurate influence estimates and degradation of future selection.

\subsection{Regularizing second-order influences}
To mitigate the harmful second-order effects without imposing additional memory overhead, we suggest regularizing each coreset selection step in advance, so that future influence estimates will be less error-prone.

Suppose we are at the $t$-th step, the goal is to minimize its interference with some sample $z'$ of the next step, namely the total effect of discarding the subset $\overline{\mathcal{C}}_t=\mathcal{C}_{t-1}\cup\mathcal{Z}_t-\mathcal{C}_t$, which can be computed by summing~\cref{eq:if_second_1} or~\cref{eq:if_second_2}\footnote{With first-order approximations here, the group effect~\cite{basu2020second} of dropped datapoints can be fairly neglected.}. However, it cannot be predetermined which equation to be used, so a hyperparameter $\mu\in[0,1]$ is introduced as the ratio for the latter case, allowing us to consider a weighted sum of the two second-order influences:
\begin{equation}
\label{eq:if_delta}
\resizebox{\linewidth}{!}{$
\begin{aligned}
&\Delta\mathcal{I}(z')\approx-\sum_{z\in\overline{\mathcal{C}}_t}\mathcal{I}^{(2)}(z,z')\cdot1\\
&=\sum_{z\in\overline{\mathcal{C}}_t}(\nabla_\theta L(z,\hat\theta_{t+1})-\mu H_{\hat\theta_{t+1},z}s_{t+1})^TH_{\hat\theta_{t+1}}^{-1}\nabla_\theta L(z',\hat\theta_{t+1}).
\end{aligned}
$}
\end{equation}

Ideally, the magnitude of total influence $\Delta I(z')$ should be very small so that future selection will receive less interference. However, this term is intractable since it is associated with an unknown sample $z'$, so we turn to optimize its upper bound given by the Cauchy–Schwarz inequality:
\begin{equation}
\begin{aligned}
|\Delta\mathcal{I}(z')|\le&\Biggl\|\sum_{z\in\overline{\mathcal{C}}_t}(\nabla_\theta L(z,\hat\theta_{t+1})-\mu H_{\hat\theta_{t+1},z}s_{t+1})\Biggr\|\\&\times\Big\|H_{\hat\theta_{t+1}}^{-1}\nabla_\theta L(z',\hat\theta_{t+1})\Big\|,
\end{aligned}
\end{equation}
We omit the second norm that depends mainly on the upcoming data, and employ the first one to regularize the ongoing selection, during which the yet unknown $\theta_{t+1}$ and $s_{t+1}$ are approximated with $\theta_t$ and $s_t$, respectively:
\begin{equation}
\label{eq:reg}
\mathcal{R}(\mathcal{C}_t)=\Biggl\|\sum_{z\in\overline{\mathcal{C}}_t}(\nabla_\theta L(z,\hat\theta_t)-\mu H_{\hat\theta_t,z}s_t)\Biggr\|.
\end{equation}

\paragraph{Proposed selection criterion.} A weighted sum of the first-order influence and second-order regularizer balanced by a hyperparameter $\nu$ is adopted to guide the selection process. The final objective at the $t$-th step is as follows:
\begin{equation}
\label{eq:criterion}
\min_{\mathcal{C}_t\subset\mathcal{C}_{t-1}\cup\mathcal{Z}_t,|\mathcal{C}_t|\le m }\sum_{z\in\mathcal{C}_t}\mathcal{I}(z)+\nu\mathcal{R}(\mathcal{C}_t).
\end{equation}
Next, we will interpret it by unveiling its connection to two widely adopted selection criteria, and then present an efficient implementation for deep models in practical use.

\begin{figure}
    \centering
    \includegraphics[width=\linewidth]{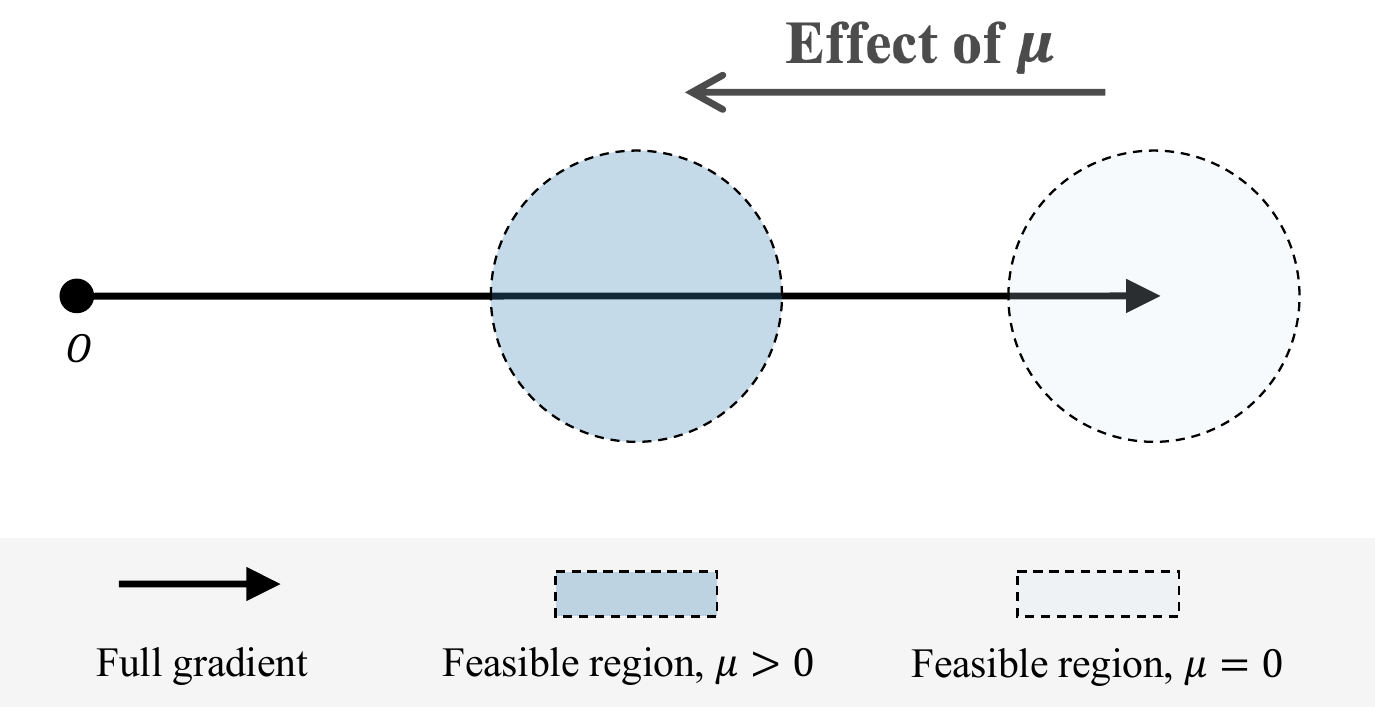}
    \caption{Geometric interpretation of our proposed regularizer. As $\mu$ grows larger from zero, the regularizer shifts from the initial gradient matching to introducing more gradient diversity along the predominant direction.}
    \label{fig:interpret}
\end{figure}

\subsection{Connection to gradient matching and diversity}
\label{sect:connection}

The following case studies of~\cref{eq:reg} allow to establish clear connections between the proposed regularizer and two commonly used selection criteria: gradient matching~\cite{zhao2021dataset, killamsetty2021grad, balles2021gradient} and sample diversity~\cite{aljundi2019gradient, bang2021rainbow, yoon2022online}.

Start by setting $\mu=0$, which corresponds to having full confidence that the currently discarded samples will not be favored in the future, and then the regularizer becomes:
\begin{equation}
\mathcal{R}(\mathcal{C}_t)=\Biggl\|\smashoperator[r]{\sum_{z\in\mathcal{C}_{t-1}\cup\mathcal{Z}_t}}\;\nabla_\theta L(z,\hat\theta_t)-\sum_{z\in\mathcal{C}_t}\nabla_\theta L(z,\hat\theta_t)\Biggr\|.
\end{equation}
It turns out to minimize the Euclidean distance between the full gradient and the coreset gradient, which is equivalent to gradient matching. For an intuitive look, we consider a hard constraint $\mathcal{R}(\mathcal{C}_t)<\delta$ and mark the feasible region of the coreset gradient with light blue in~\cref{fig:interpret}.

In the more general case $\mu>0$, we temporarily ignore the sample information carried in Hessian for simplicity, by assuming that the Hessian is identical for all training samples. Then there is:
\begin{equation}
\mathcal{R}(\mathcal{C}_t)=\Biggl\|(1-\alpha\mu)\;\smashoperator[lr]{\sum_{z\in\mathcal{C}_{t-1}\cup\mathcal{Z}_t}}\;\nabla_\theta L(z,\hat\theta_t)-\sum_{z\in\mathcal{C}_t}\nabla_\theta L(z,\hat\theta_t)\Biggr\|,
\end{equation}
where $\alpha$ is a coefficient related to the coreset size only. As illustrated in~\cref{fig:interpret}, the feasible region is moved in the opposite direction of the total gradient. This promotes samples whose gradients are less aligned or even conflicting with the main direction to be included in the coreset, thus encouraging diversity. In a holistic view, the hyperparameter $\mu$ plays the role of balancing gradient matching and diversity.

While these simplifications equate our proposed regularizer to a couple of intuitive criteria, the original form excels in the additional incorporation of Hessian-related information, which will be empirically demonstrated in the experiments.

\subsection{Implementation for neural networks}
\label{sect:implementaton}
This section presents an efficient implementation of optimizing the selection criterion with deep neural networks. Since similar issues for the vanilla influence functions have been intensively addressed~\cite{koh2017understanding, borsos2020coresets, sun2022exploring}, the focus here is on the optimization of the newly proposed regularizer.

A key difference of our second-order regularizer is that it cannot be partitioned into independent terms associated to different samples, which hinders application of the aforementioned greedy selection strategy. We overcome this challenge by leveraging the first-order Taylor expansion of the regularizer. To this end, the original variable $\mathcal{C}_t$ is replaced by a continuous vector $w_t$, where each element $w_{t,i}=\mathds{1}(z_i\in\mathcal{C}_t)$ indicates whether the corresponding sample is included in the coreset. After that, the regularizer can be transformed into an equivalent form:
\begin{equation}
\label{eq:reg_new}
\mathcal{R}(w_t)=\Biggl\|\smashoperator[r]{\sum_{z_i\in\mathcal{C}_{t-1}\cup\mathcal{Z}_t}}\;(1-w_{t,i})(\nabla_\theta L(z_i,\hat\theta_t)-\mu H_{\hat\theta_t,z_i}s_t)\Biggr\|,
\end{equation}
whose first-order Taylor approximation \wrt $w_t$ is a linear combination of uncorrelated terms (see Supplementary for details) and can therefore be addressed with greedy heuristics. We then implement an iterative optimization algorithm for selection that starts from an exterior point $w_t=\mathbf{1}$, and modifies $w_t$ during each iteration to drop the sample $z_i$ with the largest $I(z_i)+\nu\nabla_{w_{t,i}}\mathcal{R}(w_t)$, until the memory constraint $\|w_t\|_0\le m$ is satisfied. Such a strategy of greedily minimizing the proposed criterion is found to be sufficient in later experiments.

To further enable efficient inference on non-convex and over-parameterized deep neural networks, we are inspired by Borsos \etal~\cite{borsos2020coresets} and employ neural tangent kernels~\cite{jacot2018neural} as the proxy model for computing the regularizer in~\cref{eq:reg_new}. The inverse Hessian-vector product $s_t$ therein can be effectively computed using the conjugate gradient method, with a small damping term added to the Hessian to ensure that it is positive definite.

\paragraph{Computational cost.} The performance bottleneck lies in computing the inverse Hessian-vector product $s_t$, for which the conjugate gradient method takes $O(m'p^2)$ time to yield an exact solution~\cite{koh2017understanding}, where $m'=|\mathcal{C}_{t-1}\cup\mathcal{Z}_t|$ and $p$ is the number of parameters. Since this part of cost is shared with the first-order influence function, our regularizer introduces only a small overhead of $O(m'p)$ in evaluating the Hessian-vector product $H_{\hat\theta_t,z_i}s_t$ each time. In addition, the adopted neural tangent kernel approximation significantly eases the burden by reducing the number of parameters.

Our computational efficiency will be further verified by the runtime comparison in~\cref{sect:ablation} and the per-step overhead analysis in the supplementary material.

\begin{table*}
    \centering
    \resizebox{\linewidth}{!}{
    \begin{tabular}{llcccccccc}
    \toprule
    \multicolumn{2}{l}{\multirow{3}{*}{Method}} & \multicolumn{4}{c}{Class-incremental} & \multicolumn{4}{c}{Task-incremental} \\
    \cmidrule(lr){3-6} \cmidrule(l){7-10} & & \multicolumn{2}{c}{$m=300$} & \multicolumn{2}{c}{$m=500$} & \multicolumn{2}{c}{$m=300$} & \multicolumn{2}{c}{$m=500$} \\
    \cmidrule(lr){3-4} \cmidrule(lr){5-6} \cmidrule(lr){7-8} \cmidrule(l){9-10} & & ACC (\%) & BWT (\%) & ACC (\%) & BWT (\%) & ACC (\%) & BWT (\%) & ACC (\%) & BWT (\%) \\
    \midrule
    \multirow{8}{*}{Non-IF} & GEM~\cite{lopez2017gradient} & 37.51 & -70.48 & 36.95 & -69.76 & 89.34 & -9.09 & 90.42 & -7.88 \\
    & A-GEM~\cite{chaudhry2019efficient} & 20.02 & -95.68 & 20.01 & -95.69 & 85.52 & -14.07 & 86.45 & -12.83 \\
    & ER~\cite{chaudhry2019tiny} & 34.19 & -78.18 & 40.45 & -70.36 & 88.97 & -9.95 & 90.60 & -7.74 \\
    & GSS~\cite{aljundi2019gradient} & 35.89 & -75.80 & 41.96 & -68.24 & 88.05 & -10.63 & 90.38 & -7.73 \\
    & ER-MIR~\cite{aljundi2019online} & 38.53 & -72.72 & 42.65 & -67.50 & 88.50 & -10.33 & 90.63 & -7.62 \\
    & GDUMB~\cite{prabhu2020gdumb} & 36.92 & - & 44.27 & - & 73.22 & - & 78.06 & - \\
    & HAL~\cite{chaudhry2021using} & 24.45 & -83.56 & 27.94 & -80.01 & 79.90 & -14.39 & 81.84 & -12.73 \\
    & GMED~\cite{jin2021gradient} & 38.12 & -73.16 & 43.68 & -66.21 & 88.91 & -9.76 & 89.72 & -8.75 \\
    \cmidrule(){1-10} \multirow{3}{*}{IF} & Vanilla IF & 41.76 & -68.59 & 47.14 & -62.20 & 90.67 & -7.65 & 91.06 & -7.36 \\
    & MetaSP~\cite{sun2022exploring} & 43.76 & -66.37 & 50.10 & -58.39 & 89.91 & -9.00 & 91.41 & -7.36\\
    & Ours & \textbf{48.62} & \textbf{-60.24} & \textbf{53.07} & \textbf{-54.44} & \textbf{91.52} & \textbf{-6.94} & \textbf{92.53} & \textbf{-5.46}\\
    \bottomrule
    \end{tabular}
    }
    \caption{Comparison with influence function (IF)-based methods and non-IF-based methods on Split CIFAR-10 under different memory constraints. For the compared methods, we report the results summarized in~\cite{sun2022exploring} in chronological order. The task-incremental setting asks the model to classify within each task, while the class-incremental setting requires predicting both task identity and class label.}
    \label{tab:cifar10}
\end{table*}

\section{Experiments}

\subsection{Experimental setup}
\label{sect:setup}

\paragraph{Datasets.} We conduct experiments on three continual learning benchmarks: (1) Split CIFAR-10~\cite{zenke2017continual} splits the original CIFAR-10~\cite{krizhevsky2009learning} dataset into 5 disjoint subsets, where each subset comprises 2 classes. (2) Split CIFAR-100~\cite{zenke2017continual} is constructed from the CIFAR-100~\cite{krizhevsky2009learning} dataset, containing 10 tasks with disjoint class labels. (3) Split \textit{mini}ImageNet~\cite{chaudhry2019tiny, sun2022exploring} derives from the few-shot learning dataset \textit{mini}ImageNet~\cite{vinyals2016matching}, a subset of ImageNet~\cite{deng2009imagenet} with 100 classes and 600 images per class. The dataset is divided equally to create 5 sequential tasks, with each image resized to $32\times32$. For all three benchmarks, we follow their original papers in splitting training and test sets.

\paragraph{Metrics.} Two evaluation metrics are employed, including Average Accuracy (ACC) and Backward Transfer (BWT)~\cite{lopez2017gradient}, where ACC is the average accuracy after the model has been trained on all tasks and BWT indicates the average forgetting of all previous tasks. Formally, they are defined as:
\begin{equation}
\textrm{ACC}=\frac{1}{T}\sum_{i=1}^T R_{T,i},\;\;\textrm{BWT}=\frac{1}{T-1}\sum_{i=1}^{T-1}R_{T,i}-R_{i,i},
\end{equation}
where $T$ is the number of tasks and $R_{i,j}$ is the accuracy of the model on the $j$-th task after learning $i$ tasks. As done in~\cite{buzzega2020dark, sun2022exploring}, we measure the metrics under both task-incremental and class-incremental settings. The latter is particularly difficult as it does not provide task identity for each sample at test time.

\paragraph{Baselines.} Our method is compared against nine replay-based competitors: ER~\cite{chaudhry2019tiny}, GEM~\cite{lopez2017gradient}, A-GEM~\cite{chaudhry2019efficient}, GSS~\cite{aljundi2019gradient}, ER-MIR~\cite{aljundi2019online}, GDUMB~\cite{prabhu2020gdumb}, HAL~\cite{chaudhry2021using}, GMED~\cite{jin2021gradient} and MetaSP~\cite{sun2022exploring}. Additionally, a base strategy that uses the vanilla influence functions (IF) introduced in~\cref{sect:ifs} is considered. To make a fair comparison, we first reproduce the main experiments of~\cite{sun2022exploring} and then report the results therein.

\paragraph{Implementation details.} We adopt ResNet-18~\cite{he2016deep} as the backbone architecture. The model is optimized by SGD for 50 epochs per task, with fixed batch size and replay batch size of both 32, as in~\cite{buzzega2021rethinking}. The learning rate is set to 0.1 on Split CIFAR-10 and Split CIFAR-100, and 0.03 on Split \textit{mini}ImageNet. The other hyperparameters are empirically set as $\mu=0.5$ and $\nu=0.01$ by default, whose sensitivity analysis will be given in~\cref{sect:ablation}. In the training, we use cross entropy loss for replay samples and apply standard data augmentations of random cropping and horizontal flipping. To limit the computational overhead of our method, the replay buffer is only updated during the last epoch of each task. The continual learning scenarios are provided by Mammoth~\cite{buzzega2020dark, boschini2022class} which relies on PyTorch~\cite{paszke2019pytorch}. For calculating Neural Tangent Kernels, the library of~\cite{novak2020neural} is employed.

\begin{table*}
\captionsetup{skip=0em}
\captionsetup[subtable]{belowskip=1em}
\begin{subtable}{\linewidth}
    \centering
    \resizebox{\linewidth}{!}{
    \begin{tabular}{llcccccccc}
    \toprule
    \multicolumn{2}{l}{\multirow{3}{*}{Method}} & \multicolumn{4}{c}{Class-incremental} & \multicolumn{4}{c}{Task-incremental} \\
    \cmidrule(lr){3-6} \cmidrule(l){7-10} & & \multicolumn{2}{c}{$m=500$} & \multicolumn{2}{c}{$m=1000$} & \multicolumn{2}{c}{$m=500$} & \multicolumn{2}{c}{$m=1000$} \\
    \cmidrule(lr){3-4} \cmidrule(lr){5-6} \cmidrule(lr){7-8} \cmidrule(l){9-10} & & ACC (\%) & BWT (\%) & ACC (\%) & BWT (\%) & ACC (\%) & BWT (\%) & ACC (\%) & BWT (\%) \\
    \midrule
    \multirow{8}{*}{Non-IF} & GEM~\cite{lopez2017gradient} & 15.91 & -77.07 & 22.79 & -68.32 & 68.68 & -18.72 & 73.71 & -12.81 \\
    & A-GEM~\cite{chaudhry2019efficient} & 9.31 & -85.18 & 9.27 & -84.88 & 55.28 & -34.10 & 55.95 & -33.01 \\
    & ER~\cite{chaudhry2019tiny} & 13.75 & -81.64 & 17.56 & -77.52 & 66.82 & -22.73 & 71.74 & -17.40 \\
    & GSS~\cite{aljundi2019gradient} & 14.01 & -80.02 & 17.87 & -76.04 & 66.80 & -21.44 & 71.98 & -16.06 \\
    & ER-MIR~\cite{aljundi2019online} & 13.49 & -82.09 & 17.56 & -77.59 & 66.18 & -23.60 & 71.20 & -18.10 \\
    & GDUMB~\cite{prabhu2020gdumb} & 11.11 & - & 15.75 & - & 36.40 & - & 43.25 & - \\
    & HAL~\cite{chaudhry2021using} & 8.20 & \textbf{-65.70} & 10.59 & \textbf{-63.86} & 44.98 & -25.17 & 50.07 & -20.61 \\
    & GMED~\cite{jin2021gradient} & 14.56 & -80.68 & 18.67 & -76.23 & 68.82 & -20.53 & 73.91 & -15.10 \\
    \cmidrule(){1-10} \multirow{3}{*}{IF} & Vanilla IF & 17.49 & -77.54 & 22.75 & -72.56 & 71.74 & -17.90 & 73.25 & -17.22\\
    & MetaSP~\cite{sun2022exploring} & 19.28 & -76.13 & 25.72 & -68.69 & 70.81 & -19.74 & \textbf{76.14} & \textbf{-14.32}\\
    & Ours & \textbf{21.15} & -73.24 & \textbf{27.99} & -64.56 & \textbf{72.53} & \textbf{-17.22} & 74.27 & -16.37 \\
    \bottomrule
    \end{tabular}
    }
    \caption{Split CIFAR-100}
    \label{tab:cifar100}
\end{subtable}
\begin{subtable}{\linewidth}
    \centering
    \resizebox{\linewidth}{!}{
    \begin{tabular}{llcccccccc}
    \toprule
    \multicolumn{2}{l}{\multirow{3}{*}{Method}} & \multicolumn{4}{c}{Class-incremental} & \multicolumn{4}{c}{Task-incremental} \\
    \cmidrule(lr){3-6} \cmidrule(l){7-10} & & \multicolumn{2}{c}{$m=500$} & \multicolumn{2}{c}{$m=1000$} & \multicolumn{2}{c}{$m=500$} & \multicolumn{2}{c}{$m=1000$} \\
    \cmidrule(lr){3-4} \cmidrule(lr){5-6} \cmidrule(lr){7-8} \cmidrule(l){9-10} & & ACC (\%) & BWT (\%) & ACC (\%) & BWT (\%) & ACC (\%) & BWT (\%) & ACC (\%) & BWT (\%) \\
    \midrule
    \multirow{6}{*}{Non-IF} & A-GEM~\cite{chaudhry2019efficient} & 10.69 & -49.22 & 10.69 & -49.16 & 18.34 & -39.65 & 18.78 & -39.05 \\
    & ER~\cite{chaudhry2019tiny} & 11.00 & -50.84 & 11.35 & -50.08 & 28.97 & -28.40 & 31.59 & -24.95 \\
    & GSS~\cite{aljundi2019gradient} & 11.09 & -50.66 & 11.42 & -49.91 & 28.67 & -28.71 & 31.75 & -24.56 \\
    & ER-MIR~\cite{aljundi2019online} & 11.07 & -50.46 & 11.32 & -49.92 & 29.10 & -27.95 & 31.39 & -24.89 \\
    & GDUMB~\cite{prabhu2020gdumb} & 6.22 & - & 7.15 & - & 16.37 & - & 17.69 & - \\
    & GMED~\cite{jin2021gradient} & 11.03 & -50.23 & 11.73 & -48.93 & 30.47 & -26.02 & 32.85 & -22.69 \\
    \cmidrule(){1-10} \multirow{3}{*}{IF} & Vanilla IF & 12.08 & -48.55 & 14.64 & -47.15 & 33.74 & -21.71 & 37.55 & -19.28\\
    & MetaSP~\cite{sun2022exploring} & 12.74 & -48.84 & 14.54 & -45.52 & 34.36 & -21.70 & 37.20 & -17.83\\
    & Ours & \textbf{13.63} & \textbf{-47.94} & \textbf{16.15} & \textbf{-43.78} & \textbf{36.46} & \textbf{-19.48} & \textbf{39.61} & \textbf{-16.01} \\
    \bottomrule
    \end{tabular}
    }
    \caption{Split \textit{mini}ImageNet}
    \label{tab:miniimg}
\end{subtable}
\caption{Comparison with state-of-the-art methods on more challenging benchmarks, including Split CIFAR-100 and Split \textit{mini}ImageNet. While most setups follow the previous experiment, a different hyperparameter setting $\mu=0.75$ is adopted on Split CIFAR-100 with memory size $m=1000$ empirically.}
\label{tab:larger}
\end{table*}

\subsection{Main results}

The results on the Split CIFAR-10 benchmark are summarized in~\cref{tab:cifar10}. Overall, our approach achieves the best continual learning performance in all metrics under various evaluation settings and memory constraints, suggesting its superiority over state-of-the-art methods. From the results, we also observe that: (1) The first-order influence-based methods obtain significant performance gains over traditional methods, but our version almost doubles the improvement under the small memory setting of $m=300$. By taking into account the second-order influences, it outperforms the state-of-the-art method by 4.86\% in ACC and 6.13\% in BWT in the challenging class-incremental evaluations, which in turn reflects the non-trivial effect of second-order influences. (2) Though the base strategy vanilla IF is slightly inferior to the state-of-the-art method MetaSP in most cases for it does not introduce any new losses in the model training, it can substantially surpass MetaSP with a simple regularization term on the selection criterion. This verifies the effectiveness of our proposed regularizer.

We further evaluate our method on the more difficult Split CIFAR-100 and Split \textit{mini}ImageNet benchmarks. As presented in~\cref{tab:larger}, our approach brings consistent performance improvement over vanilla IF on both benchmarks, by up to 5.24\% and 8.0\% in terms of ACC and BWT, respectively. Compared to other approaches, it obtains state-of-the-art performance on Split \textit{mini}ImageNet in all metrics, and leads ACC by about 2\% on three different setups of Split CIFAR-100.  Only in the simplest scenario of task-incremental learning on Split CIFAR-100 with $m=1000$, our method fails to surpass MetaSP, for which we speculate that the high-order interference is alleviated to some extent by the large buffer itself. Nevertheless, our method remains competitive in most evaluations.

\subsection{Ablation study and analysis}
\label{sect:ablation}

This section empirically justifies the design of our approach through a series of experiments conducted on Split CIFAR-10 with a fixed memory size of $m=500$.

\paragraph{Comparison with other regularizers.} To further validate the effectiveness of our regularizer, it is compared to two closely related criteria, namely gradient matching and diversity, in~\cref{tab:regularizer}. Our regularizer draws on the strengths from both criteria, including the satisfactory task-incremental performance of gradient matching and the higher class-incremental ACC of gradient diversity. In addition, we compare it with a combination of both criteria, during which the considerable performance advantage of our Hessian-aware regularizer illustrates the benefits from incorporating Hessian-related information. More comparative studies can be found in the supplementary material.

\begin{table}
    \centering
    \resizebox{\linewidth}{!}{
    \begin{tabular}{lcccc}
    \toprule
    \multirow{2}{*}{Method} & \multicolumn{2}{c}{Class-incremental} & \multicolumn{2}{c}{Task-incremental} \\
    \cmidrule(lr){2-3} \cmidrule(l){4-5} & ACC (\%) & BWT (\%) & ACC (\%) & BWT (\%) \\
    \midrule
    Vanilla IF & 47.14 & -62.20 & 91.06 & -7.36 \\
     + matching & 49.86 & -58.24 & 92.30 & -5.78 \\
     + diversity & 50.59 & -57.06 & 90.69 & -7.28 \\
     + both & 51.61 & -56.31 & 91.93 & -6.11\\
    \cmidrule(){1-5}
    Ours & \textbf{53.07} & \textbf{-54.44} & \textbf{92.53} & \textbf{-5.46} \\
    \bottomrule
    \end{tabular}
    }
    \caption{Comparison with alternative regularizers on Split CIFAR-10 with memory size $m=500$. The diversity-based regularizer is implemented by minimizing the norm of the coreset gradient.}
    \label{tab:regularizer}
\end{table}
\begin{figure}
    \centering
    \newlength{\imagewidth}
    \settowidth{\imagewidth}{\includegraphics{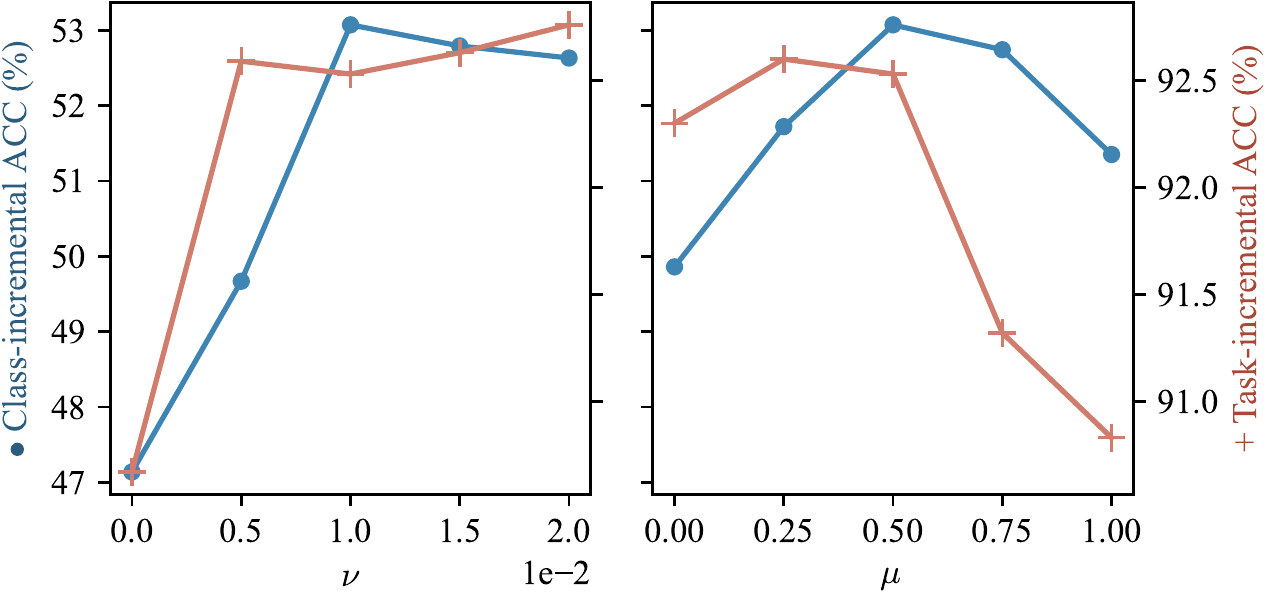}}
    \begin{subfigure}[t]{0.5\linewidth}
        \includegraphics[trim={0 0 0.5\imagewidth{} 0}, clip, width=\linewidth]{figs/hyperparameter.pdf}
        \caption{Weight for the regularizer}
        \label{fig:nu}
    \end{subfigure}\hfill
    \begin{subfigure}[t]{0.5\linewidth}
        \includegraphics[trim={0.5\imagewidth{} 0 0 0}, clip, width=\linewidth]{figs/hyperparameter.pdf}
        \caption{Weight for Hessian-related term}
        \label{fig:mu}
    \end{subfigure}
    \caption{Sensitivity to hyperparameters on Split CIFAR-10 with $m=500$. They are set to $\nu=0.01$ and $\mu=0.5$ by default.}
    \label{fig:hyperparameter}
\end{figure}

\paragraph{Hyperparameter sensitivity.} In order to explore the sensitivity of our strategy to hyperparameter settings, we vary the two coefficients $\mu$ and $\nu$ and plot the response of model performance.~\Cref{fig:nu} shows that under both evaluation settings, ACC improves consistently with the growth of weight $\nu$ on the second-order regularizer until its stability. The situation is more complicated in~\cref{fig:mu}, where the two ACC curves exhibit inconsistent trends regarding the weight $\mu$ for the Hessian-related term. Recall that in~\cref{sect:connection} $\mu$ is associated with the degree of diversity, so it may be interpreted that the more challenging class-incremental setting requires higher memory diversity, while the simpler task-increment setting accommodates lower diversity.

\begin{figure}
\begin{minipage}[b]{.46\linewidth}
    \centering
    \captionsetup{skip=0.5em}
    \includegraphics[width=\linewidth]{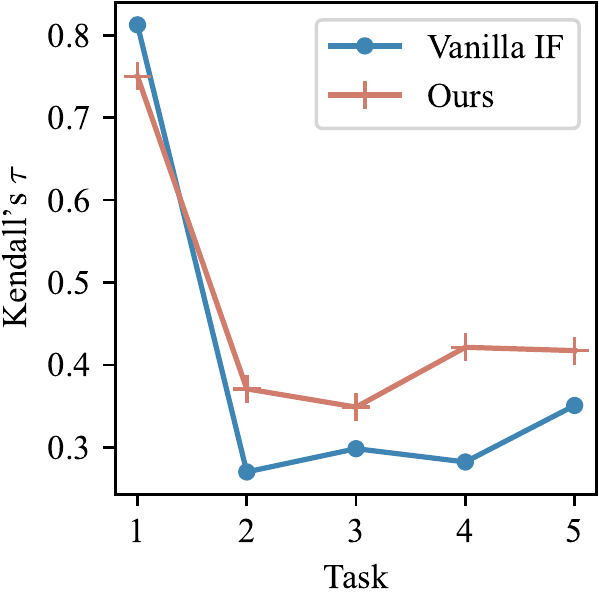}
    \caption{Evolution of the influence estimation accuracy measured by Kendall's $\tau$~\cite{kendall1938new}.}
    \label{fig:tau}
\end{minipage}\hfill
\begin{minipage}[b]{.5\linewidth}
\resizebox{\linewidth}{!}{
    \begin{tabular}{lc}
    \toprule
    Method & Runtime (h) \\
    \midrule
    ER~\cite{chaudhry2019tiny} & 0.92 \\
    A-GEM~\cite{chaudhry2019efficient} & 1.22 \\
    GSS~\cite{aljundi2019gradient} & 0.92 \\
    HAL~\cite{chaudhry2019efficient} & 1.73 \\
    \cmidrule{1-2}
    Vanilla IF & 1.69 \\
    MetaSP~\cite{sun2022exploring} & 2.03 \\
    Ours & 1.74 \\
    \bottomrule
    \end{tabular}
}
\captionof{table}{Running time measured on Split CIFAR-10 using a single NVIDIA 2080 Ti GPU. }
\label{tab:runtime}
\end{minipage}
\end{figure}

\paragraph{Accuracy of influence estimates.} To measure accuracy, the previous practice of comparing with the ground-truth given by leave-one-out retraining~\cite{koh2017understanding, basu2020second} imposes an impractical cost under the continual learning setup. Hence, we turn to a large reservoir sampling buffer to produce unbiased influence estimates and compute their correlation with other predictions. The Kendall rank correlation~\cite{kendall1938new} is adopted since we are more concerned with each sample's rank during selection. As illustrated in~\cref{fig:tau}, the buffer maintained by vanilla IF dissociates from the unbiased buffer after a short period, producing inaccurate influence estimates. In contrast, our strategy yields a relatively higher agreement, which verifies its effectiveness in mitigating interference.

\paragraph{Running time.}~\Cref{tab:runtime} compares our proposed method with several replay-based competitors in terms of efficiency. Though influence-based approaches generally incur higher time costs in the Hessian-related computations, our method imposes a limited overhead with the newly introduced second-order regularizer. This implies that the above improvements are achieved within comparable training time, demonstrating the high efficiency of our implementation.

\section{Concluding remarks}

We propose an effective coreset selection strategy for continual learning that addresses the interference within successive selection steps. By dissecting the interactions between consecutive rounds of influence-based selection, a new class of second-order influences is identified, with which the long selection process may gradually lose diversity and accumulate bias. To solve this problem, a novel regularizer is presented which is linked to two other popular criteria but incorporates additional Hessian-related information. Finally, we implement the proposed selection criterion with high efficiency and validate its effectiveness in a variety of comparative experiments.

{\footnotesize This work is supported by National Key R\&D Program of China (2020AAA0104401), Beijing Natural Science Foundation (Z190001) and Peng Cheng Laboratory Key Research Project No.PCL2021A07.}

{\small
\bibliographystyle{ieee_fullname}
\bibliography{main}

\begin{thebibliography}{10}\itemsep=-1pt

\bibitem{aljundi2019online}
Rahaf Aljundi, Eugene Belilovsky, Tinne Tuytelaars, Laurent Charlin, Massimo
  Caccia, Min Lin, and Lucas Page-Caccia.
\newblock Online continual learning with maximal interfered retrieval.
\newblock In {\em NeurIPS}, pages 11849--11860, 2019.

\bibitem{aljundi2017expert}
Rahaf Aljundi, Punarjay Chakravarty, and Tinne Tuytelaars.
\newblock Expert gate: Lifelong learning with a network of experts.
\newblock In {\em CVPR}, pages 3366--3375, 2017.

\bibitem{aljundi2019gradient}
Rahaf Aljundi, Min Lin, Baptiste Goujaud, and Yoshua Bengio.
\newblock Gradient based sample selection for online continual learning.
\newblock In {\em NeurIPS}, pages 11817--11826, 2019.

\bibitem{balles2021gradient}
Lukas Balles, Giovanni Zappella, and C{\'e}dric Archambeau.
\newblock Gradient-matching coresets for continual learning.
\newblock In {\em NeurIPS Workshops}, 2021.

\bibitem{bang2021rainbow}
Jihwan Bang, Heesu Kim, YoungJoon Yoo, Jung-Woo Ha, and Jonghyun Choi.
\newblock Rainbow memory: Continual learning with a memory of diverse samples.
\newblock In {\em CVPR}, pages 8218--8227, 2021.

\bibitem{basu2020second}
Samyadeep Basu, Xuchen You, and Soheil Feizi.
\newblock On second-order group influence functions for black-box predictions.
\newblock In {\em ICML}, pages 715--724, 2020.

\bibitem{borsos2020coresets}
Zal{\'a}n Borsos, Mojmir Mutny, and Andreas Krause.
\newblock Coresets via bilevel optimization for continual learning and
  streaming.
\newblock In {\em NeurIPS}, pages 14879--14890, 2020.

\bibitem{boschini2022class}
Matteo Boschini, Lorenzo Bonicelli, Pietro Buzzega, Angelo Porrello, and Simone
  Calderara.
\newblock Class-incremental continual learning into the extended {DER}-verse.
\newblock {\em TPAMI}, pages 1--16, 2022.

\bibitem{buzzega2020dark}
Pietro Buzzega, Matteo Boschini, Angelo Porrello, Davide Abati, and Simone
  Calderara.
\newblock Dark experience for general continual learning: a strong, simple
  baseline.
\newblock In {\em NeurIPS}, pages 15920--15930, 2020.

\bibitem{buzzega2021rethinking}
Pietro Buzzega, Matteo Boschini, Angelo Porrello, and Simone Calderara.
\newblock Rethinking experience replay: a bag of tricks for continual learning.
\newblock In {\em ICPR}, pages 2180--2187, 2021.

\bibitem{caccia2022new}
Lucas Caccia, Rahaf Aljundi, Nader Asadi, Tinne Tuytelaars, Joelle Pineau, and
  Eugene Belilovsky.
\newblock New insights on reducing abrupt representation change in online
  continual learning.
\newblock In {\em ICLR}, 2022.

\bibitem{chaudhry2021using}
Arslan Chaudhry, Albert Gordo, Puneet Dokania, Philip Torr, and David
  Lopez-Paz.
\newblock Using hindsight to anchor past knowledge in continual learning.
\newblock In {\em AAAI}, pages 6993--7001, 2021.

\bibitem{chaudhry2019efficient}
Arslan Chaudhry, Marc’Aurelio Ranzato, Marcus Rohrbach, and Mohamed
  Elhoseiny.
\newblock Efficient lifelong learning with a-gem.
\newblock In {\em ICLR}, 2019.

\bibitem{chaudhry2019tiny}
Arslan Chaudhry, Marcus Rohrbach, Mohamed Elhoseiny, Thalaiyasingam Ajanthan,
  Puneet~K Dokania, Philip~HS Torr, and Marc'Aurelio Ranzato.
\newblock On tiny episodic memories in continual learning.
\newblock In {\em ICML Workshops}, 2019.

\bibitem{chen2020multi}
Hongge Chen, Si Si, Yang Li, Ciprian Chelba, Sanjiv Kumar, Duane Boning, and
  Cho-Jui Hsieh.
\newblock Multi-stage influence function.
\newblock In {\em NeurIPS}, pages 12732--12742, 2020.

\bibitem{chen2018lifelong}
Zhiyuan Chen and Bing Liu.
\newblock {\em Lifelong Machine Learning}.
\newblock Synthesis Lectures on Artificial Intelligence and Machine Learning.
  Morgan \& Claypool Publishers, 2018.

\bibitem{cook1982residuals}
R~Dennis Cook and Sanford Weisberg.
\newblock {\em Residuals and influence in regression}.
\newblock New York: Chapman and Hall, 1982.

\bibitem{de2021continual}
Matthias De~Lange, Rahaf Aljundi, Marc Masana, Sarah Parisot, Xu Jia,
  Ale{\v{s}} Leonardis, Gregory Slabaugh, and Tinne Tuytelaars.
\newblock A continual learning survey: Defying forgetting in classification
  tasks.
\newblock {\em TPAMI}, 44(7):3366--3385, 2021.

\bibitem{deng2009imagenet}
Jia Deng, Wei Dong, Richard Socher, Li-Jia Li, Kai Li, and Li Fei-Fei.
\newblock {ImageNet}: A large-scale hierarchical image database.
\newblock In {\em CVPR}, pages 248--255, 2009.

\bibitem{guo2021fastif}
Han Guo, Nazneen Rajani, Peter Hase, Mohit Bansal, and Caiming Xiong.
\newblock {FastIF}: Scalable influence functions for efficient model
  interpretation and debugging.
\newblock In {\em EMNLP}, pages 10333--10350, 2021.

\bibitem{hadsell2020embracing}
Raia Hadsell, Dushyant Rao, Andrei~A Rusu, and Razvan Pascanu.
\newblock Embracing change: Continual learning in deep neural networks.
\newblock {\em Trends in Cognitive Sciences}, 24(12):1028--1040, 2020.

\bibitem{hampel1974influence}
Frank~R Hampel.
\newblock The influence curve and its role in robust estimation.
\newblock {\em Journal of the American Statistical Association},
  69(346):383--393, 1974.

\bibitem{he2016deep}
Kaiming He, Xiangyu Zhang, Shaoqing Ren, and Jian Sun.
\newblock Deep residual learning for image recognition.
\newblock In {\em CVPR}, pages 770--778, 2016.

\bibitem{jacot2018neural}
Arthur Jacot, Franck Gabriel, and Cl{\'e}ment Hongler.
\newblock Neural tangent kernel: convergence and generalization in neural
  networks.
\newblock In {\em NeurIPS}, pages 8580--8589, 2018.

\bibitem{jin2021gradient}
Xisen Jin, Arka Sadhu, Junyi Du, and Xiang Ren.
\newblock Gradient-based editing of memory examples for online task-free
  continual learning.
\newblock In {\em NeurIPS}, pages 29193--29205, 2021.

\bibitem{kendall1938new}
Maurice~G Kendall.
\newblock A new measure of rank correlation.
\newblock {\em Biometrika}, 30(1/2):81--93, 1938.

\bibitem{killamsetty2021grad}
Krishnateja Killamsetty, S Durga, Ganesh Ramakrishnan, Abir De, and Rishabh
  Iyer.
\newblock Grad-match: Gradient matching based data subset selection for
  efficient deep model training.
\newblock In {\em ICML}, pages 5464--5474, 2021.

\bibitem{kirkpatrick2017overcoming}
James Kirkpatrick, Razvan Pascanu, Neil Rabinowitz, Joel Veness, Guillaume
  Desjardins, Andrei~A Rusu, Kieran Milan, John Quan, Tiago Ramalho, Agnieszka
  Grabska-Barwinska, et~al.
\newblock Overcoming catastrophic forgetting in neural networks.
\newblock {\em Proceedings of the National Academy of Sciences},
  114(13):3521--3526, 2017.

\bibitem{koh2019accuracy}
Pang~Wei Koh, Kai-Siang Ang, Hubert~HK Teo, and Percy Liang.
\newblock On the accuracy of influence functions for measuring group effects.
\newblock In {\em NeurIPS}, pages 5254--5264, 2019.

\bibitem{koh2017understanding}
Pang~Wei Koh and Percy Liang.
\newblock Understanding black-box predictions via influence functions.
\newblock In {\em ICML}, pages 1885--1894, 2017.

\bibitem{kong2022resolving}
Shuming Kong, Yanyan Shen, and Linpeng Huang.
\newblock Resolving training biases via influence-based data relabeling.
\newblock In {\em ICLR}, 2022.

\bibitem{krizhevsky2009learning}
Alex Krizhevsky.
\newblock Learning multiple layers of features from tiny images.
\newblock {\em Master's thesis, University of Toronto}, 2009.

\bibitem{li2017learning}
Zhizhong Li and Derek Hoiem.
\newblock Learning without forgetting.
\newblock {\em TPAMI}, 40(12):2935--2947, 2017.

\bibitem{lopez2017gradient}
David Lopez-Paz and Marc'Aurelio Ranzato.
\newblock Gradient episodic memory for continual learning.
\newblock In {\em NeurIPS}, page 6470–6479, 2017.

\bibitem{mccloskey1989catastrophic}
Michael McCloskey and Neal~J Cohen.
\newblock Catastrophic interference in connectionist networks: The sequential
  learning problem.
\newblock {\em Psychology of Learning and Motivation}, 24:109--165, 1989.

\bibitem{mirzadeh2020understanding}
Seyed~Iman Mirzadeh, Mehrdad Farajtabar, Razvan Pascanu, and Hassan
  Ghasemzadeh.
\newblock Understanding the role of training regimes in continual learning.
\newblock In {\em NeurIPS}, pages 7308--7320, 2020.

\bibitem{novak2020neural}
Roman Novak, Lechao Xiao, Jiri Hron, Jaehoon Lee, Alexander~A Alemi, Jascha
  Sohl-Dickstein, and Samuel~S Schoenholz.
\newblock Neural tangents: Fast and easy infinite neural networks in python.
\newblock In {\em ICLR}, 2020.

\bibitem{park2021influence}
Seulki Park, Jongin Lim, Younghan Jeon, and Jin~Young Choi.
\newblock Influence-balanced loss for imbalanced visual classification.
\newblock In {\em ICCV}, pages 735--744, 2021.

\bibitem{paszke2019pytorch}
Adam Paszke, Sam Gross, Francisco Massa, Adam Lerer, James Bradbury, Gregory
  Chanan, Trevor Killeen, Zeming Lin, Natalia Gimelshein, Luca Antiga, et~al.
\newblock {PyTorch}: An imperative style, high-performance deep learning
  library.
\newblock In {\em NeurIPS}, pages 8026--8037, 2019.

\bibitem{prabhu2020gdumb}
Ameya Prabhu, Philip~HS Torr, and Puneet~K Dokania.
\newblock {GDumb}: A simple approach that questions our progress in continual
  learning.
\newblock In {\em ECCV}, pages 524--540, 2020.

\bibitem{rebuffi2017icarl}
Sylvestre-Alvise Rebuffi, Alexander Kolesnikov, Georg Sperl, and Christoph~H
  Lampert.
\newblock {iCaRL}: Incremental classifier and representation learning.
\newblock In {\em CVPR}, pages 2001--2010, 2017.

\bibitem{riemer2019learning}
Matthew Riemer, Ignacio Cases, Robert Ajemian, Miao Liu, Irina Rish, Yuhai Tu,
  and Gerald Tesauro.
\newblock Learning to learn without forgetting by maximizing transfer and
  minimizing interference.
\newblock In {\em ICLR}, 2019.

\bibitem{rusu2016progressive}
Andrei~A Rusu, Neil~C Rabinowitz, Guillaume Desjardins, Hubert Soyer, James
  Kirkpatrick, Koray Kavukcuoglu, Razvan Pascanu, and Raia Hadsell.
\newblock Progressive neural networks.
\newblock {\em arXiv preprint arXiv:1606.04671}, 2016.

\bibitem{schioppa2022scaling}
Andrea Schioppa, Polina Zablotskaia, David Vilar, and Artem Sokolov.
\newblock Scaling up influence functions.
\newblock In {\em AAAI}, pages 8179--8186, 2022.

\bibitem{serra2018overcoming}
Joan Serra, Didac Suris, Marius Miron, and Alexandros Karatzoglou.
\newblock Overcoming catastrophic forgetting with hard attention to the task.
\newblock In {\em ICML}, pages 4548--4557, 2018.

\bibitem{shin2017continual}
Hanul Shin, Jung~Kwon Lee, Jaehong Kim, and Jiwon Kim.
\newblock Continual learning with deep generative replay.
\newblock In {\em NeurIPS}, page 2994–3003, 2017.

\bibitem{sun2022exploring}
Qing Sun, Fan Lyu, Fanhua Shang, Wei Feng, and Liang Wan.
\newblock Exploring example influence in continual learning.
\newblock In {\em NeurIPS}, 2022.

\bibitem{ting2018optimal}
Daniel Ting and Eric Brochu.
\newblock Optimal subsampling with influence functions.
\newblock In {\em NeurIPS}, pages 3654--3663, 2018.

\bibitem{tiwari2022gcr}
Rishabh Tiwari, Krishnateja Killamsetty, Rishabh Iyer, and Pradeep Shenoy.
\newblock {GCR}: Gradient coreset based replay buffer selection for continual
  learning.
\newblock In {\em CVPR}, pages 99--108, 2022.

\bibitem{vinyals2016matching}
Oriol Vinyals, Charles Blundell, Timothy Lillicrap, Daan Wierstra, et~al.
\newblock Matching networks for one shot learning.
\newblock In {\em NeurIPS}, pages 3637--3645, 2016.

\bibitem{vitter1985random}
Jeffrey~S Vitter.
\newblock Random sampling with a reservoir.
\newblock {\em ACM Transactions on Mathematical Software}, 11(1):37--57, 1985.

\bibitem{wang2018data}
Tianyang Wang, Jun Huan, and Bo Li.
\newblock Data dropout: Optimizing training data for convolutional neural
  networks.
\newblock In {\em IEEE International Conference on Tools with Artificial
  Intelligence}, pages 39--46, 2018.

\bibitem{wang2020less}
Zifeng Wang, Hong Zhu, Zhenhua Dong, Xiuqiang He, and Shao-Lun Huang.
\newblock Less is better: Unweighted data subsampling via influence function.
\newblock In {\em AAAI}, pages 6340--6347, 2020.

\bibitem{wu2019large}
Yue Wu, Yinpeng Chen, Lijuan Wang, Yuancheng Ye, Zicheng Liu, Yandong Guo, and
  Yun Fu.
\newblock Large scale incremental learning.
\newblock In {\em CVPR}, pages 374--382, 2019.

\bibitem{yoon2022online}
Jaehong Yoon, Divyam Madaan, Eunho Yang, and Sung~Ju Hwang.
\newblock Online coreset selection for rehearsal-based continual learning.
\newblock In {\em ICLR}, 2022.

\bibitem{yoon2018lifelong}
Jaehong Yoon, Eunho Yang, Jeongtae Lee, and Sung~Ju Hwang.
\newblock Lifelong learning with dynamically expandable networks.
\newblock In {\em ICLR}, 2018.

\bibitem{zeng2019continual}
Guanxiong Zeng, Yang Chen, Bo Cui, and Shan Yu.
\newblock Continual learning of context-dependent processing in neural
  networks.
\newblock {\em Nature Machine Intelligence}, 1(8):364--372, 2019.

\bibitem{zenke2017continual}
Friedemann Zenke, Ben Poole, and Surya Ganguli.
\newblock Continual learning through synaptic intelligence.
\newblock In {\em ICML}, pages 3987--3995, 2017.

\bibitem{zhang2022rethinking}
Rui Zhang and Shihua Zhang.
\newblock Rethinking influence functions of neural networks in the
  over-parameterized regime.
\newblock In {\em AAAI}, pages 9082--9090, 2022.

\bibitem{zhao2021dataset}
Bo Zhao, Konda~Reddy Mopuri, and Hakan Bilen.
\newblock Dataset condensation with gradient matching.
\newblock In {\em ICLR}, 2021.

\bibitem{zhou2021overcoming}
Fan Zhou and Chengtai Cao.
\newblock Overcoming catastrophic forgetting in graph neural networks with
  experience replay.
\newblock In {\em AAAI}, pages 4714--4722, 2021.

\end{thebibliography}
}

\newpage
\appendix

\section{Notation}
\Cref{tab:notation} summarizes the used notation for quick lookup.

\section{Continual learning framework}
The pseudocode for our learning procedure is presented in~\cref{alg:ours}. Following ER~\cite{chaudhry2019tiny}, the model is trained on a mini-batch composed of the current task data and replay examples at each time step. Meanwhile, to reduce the computational cost imposed by the selection algorithm, the replay buffer is updated only in the last epoch of each task. For the settings of hyperparameters, please refer to~\cref{sect:setup}.

\afterpage{
\begin{table}
    \centering
    \resizebox{\linewidth}{!}{
    \begin{tabular}{ll}
    \toprule
    Symbol & Description \\
    \midrule
    $\mathcal{Z}_t$ & Available data at the $t$-th step\\
    $\mathcal{Z}_{1:t}$ & Seen data till the $t$-th step \\
    $\mathcal{C}_t$ & Coreset at the $t$-th step \\
    $m$ & Maximum coreset size\\
    $L(z,\theta)$ & Loss of parameter $\theta$ on sample $z$\\
    $\hat\theta_t$ & Optimal point at the $t$-th step \\
    $\hat\theta_{\epsilon,z}$ & Optimal point after $z$ is upweighted by $\epsilon$ \\
    $H_{\hat\theta_t}$ & Hessian of $\hat\theta_t$ on coreset $\mathcal{C}_t$\\
    $H_{\hat\theta_t,z}$ & Hessian of $\hat\theta_t$ on sample $z$ \\
    $s_t$ & Inverse Hessian-vector product at the $t$-th step\\
    $\mathcal{I}(z)$ & Influence of $z$ on the test loss \\
    $\mathcal{I}_{\epsilon,z}(z')$ & Influence of $z'$ after $z$ is upweighted by $\epsilon$ \\
    $\mathcal{I}^{(2)}(z,z')$ & Second-order influence of $z$ and $z'$ \\
    $\Delta I(z')$ & Total interference on the influence of $z'$\\
    $\mathcal{R}(\cdot)$ & Our proposed regularizer \\
    \bottomrule
    \end{tabular}
    }
    \caption{Notation in the main paper.}
    \label{tab:notation}
\end{table}
\begin{algorithm}[t]
\small
\caption{Learning Procedure for Task $T$}
\label{alg:ours}
\begin{algorithmic}[1]
    \State {\textbf{Input:} Dataset $\mathcal{Z}$ of task $T$, coreset $\mathcal{C}_{t-1}$ from the last round of selection, the number of epochs $e_{\textrm{max}}$, model parameter $\theta$, learning rate $\eta$.}
    \For{$e = 1$ {\bf to} $e_{\textrm{max}}$}
        \For {each batch $\mathcal{Z}_t\in\mathcal{Z}$}
        \State{Sample a replay batch $\mathcal{B}_{\mathcal{C}}\in\mathcal{C}_{t-1}$}
        \State{$\theta \gets \theta - \eta \nabla_\theta\sum_{z\in \mathcal{Z}_t\cup\mathcal{B}_{\mathcal{C}}} L(z,\theta)$}
        \If{$e=e_{\textrm{max}}$}
            \State{Update coreset $\mathcal{C}_t\in\mathcal{C}_{t-1}\cup\mathcal{Z}_t$ by~\cref{sect:implementaton}}
            \State{$t\gets t+1$}
        \EndIf
        \EndFor
    \EndFor
\end{algorithmic}
\end{algorithm}
}

\section{Derivation of influence functions}
As a background introduction, this section provides the derivation of the first-order influence score $\mathcal{I}(z)$ in~\cref{eq:if_first}, following the idea by Koh and Liang~\cite{koh2017understanding}.

It begins with upweighting an interested sample $z$ by an infinitesimal amount $\epsilon$, after which the perturbed optimal point $\hat\theta_{\epsilon,z}$ can be written as follows:
\begin{equation}
\small
\hat\theta_{\epsilon,z}=\argmin_{\theta}\sum_{z_i\in\mathcal{C}_t}L(z_i,\theta)+\epsilon L(z,\theta).
\end{equation}
Its first-order optimality condition states that:
\begin{equation}
\small
0=\sum_{z_i\in\mathcal{C}_t}\nabla_\theta L(z_i,\hat\theta_{\epsilon,z})+\epsilon\nabla_\theta L(z,\hat\theta_{\epsilon,z}).
\end{equation}
To exploit the known optimal point $\hat\theta_t$, we apply the first-order Taylor expansion on the right-hand side:
\begin{equation}
\small
\begin{aligned}
0\approx&\biggl[\sum_{z_i\in\mathcal{C}_t}\nabla_\theta L(z_i,\hat\theta_t)+\epsilon\nabla_\theta L(z,\hat\theta_t)\biggr]\\
&+\biggl[\sum_{z_i\in\mathcal{C}_t}\nabla_\theta^2 L(z_i,\hat\theta_t)+\epsilon\nabla_\theta^2 L(z,\hat\theta_t)\biggr](\hat\theta_{\epsilon,z}-\hat\theta_t),
\end{aligned}
\end{equation}
where $o(\|\hat\theta_{\epsilon,z}-\hat\theta_t\|)$ terms are dropped. It is also assumed that $L$ is twice-differentiable and convex in $\theta$. Using the optimality condition $\sum_{z_i\in\mathcal{C}_t}\nabla_\theta L(z_i,\hat\theta_t)=0$ and the notation $H_{\hat\theta_t}=\sum_{z_i\in\mathcal{C}_t}\nabla_\theta^2 L(z_i,\hat\theta_t)$, it can be simplified to:
\begin{equation}
\small
\hat\theta_{\epsilon,z}-\hat\theta_t\approx H_{\hat\theta_t}^{-1}\nabla_\theta L(z,\hat\theta_t)\epsilon,
\end{equation}
where $o(\epsilon)$ terms are neglected. This yields the derivate of $\hat\theta_{\epsilon,z}$ \wrt $\epsilon$:
\begin{equation}
\small
\frac{d\hat\theta_{\epsilon,z}}{d\epsilon}\Big|_{\epsilon=0}=-H_{\hat\theta_t}^{-1}\nabla_\theta L(z,\hat\theta_t).
\end{equation}
Finally, the influence of a particular sample $z$ on the test loss can be computed by the chain rule:
\begin{equation}
\small
\begin{aligned}
\mathcal{I}(z)&=\smashoperator[r]{\sum_{z_i\in\mathcal{C}_{t-1}\cup\mathcal{Z}_t}}\;\;\frac{dL(z_i,\hat\theta_{\epsilon,z})}{d\epsilon}\Big|_{\epsilon=0}\\
&=\smashoperator[r]{\sum_{z_i\in\mathcal{C}_{t-1}\cup\mathcal{Z}_t}}\;\;\nabla_\theta L(z_i,\hat\theta_t)^\top \frac{d\hat\theta_{\epsilon,z}}{d\epsilon}\Big|_{\epsilon=0}\\
&=-\;\;\smashoperator[lr]{\sum_{z_i\in\mathcal{C}_{t-1}\cup\mathcal{Z}_t}}\;\;\nabla_\theta L(z_i,\hat\theta_t)^\top H_{\hat\theta_t}^{-1}\nabla_\theta L(z,\hat\theta_t).
\end{aligned}
\end{equation}

\section{Derivation of the second-order influence}
This section explains the derivation of the second-order effects $\mathcal{I}^{(2)}(z,z')$ in~\cref{eq:if_second_2} of~\cref{sect:ifs_second}. The derivation applies to~\cref{eq:if_second_1} as well, since they share a similar form.

In that case, the influence score of a subsequent sample $z'$ after the previous $z$ is upweighted by $\epsilon$ is as follows:
\begin{equation}
\label{eq:if_perturbed}
\small
\begin{aligned}
\mathcal{I}_{\epsilon,z}(z')&=-\biggl(\smashoperator[r]{\sum_{z_i\in\mathcal{C}_t\cup\mathcal{Z}_{t+1}}}\;\nabla_\theta L(z_i,\hat\theta_{t+1})+\epsilon\nabla_\theta L(z,\hat\theta_{t+1})\biggr)^\top\\
&\qquad\Bigl(H_{\hat\theta_{t+1}}+\epsilon H_{\hat\theta_{t+1},z}\Bigr)^{-1}\nabla_\theta L(z',\hat\theta_{t+1}).
\end{aligned}
\end{equation}
The inverse matrix therein can be effectively approximated with a Neumann series as $\epsilon\rightarrow 0$:
\begin{equation}
\small
\begin{aligned}
(A+\epsilon B)^{-1}&=A^{-1}(I+\epsilon BA^{-1})^{-1}\\
&=A^{-1}\sum_{k=0}^{\infty}(-\epsilon BA^{-1})^k\\
&=A^{-1}-\epsilon A^{-1}BA^{-1}+o(\epsilon).
\end{aligned}
\end{equation}
Take $A=H_{\hat\theta_{t+1}}$ and $B=H_{\hat\theta_{t+1},z}$ and substitute into \cref{eq:if_perturbed}, then we get:
\begin{equation}
\resizebox{\linewidth}{!}{$
\begin{aligned}
\mathcal{I}_{\epsilon,z}(z')&=-\biggl(\smashoperator[r]{\sum_{z_i\in\mathcal{C}_t\cup\mathcal{Z}_{t+1}}}\;\nabla_\theta L(z_i,\hat\theta_{t+1})+\epsilon\nabla_\theta L(z,\hat\theta_{t+1})\biggr)^\top\\
&\qquad\Bigl(H_{\hat\theta_{t+1}}^{-1}-\epsilon H_{\hat\theta_{t+1}}^{-1}H_{\hat\theta_{t+1},z}H_{\hat\theta_{t+1}}^{-1}+o(\epsilon)\Bigr)\nabla_\theta L(z',\hat\theta_{t+1}),
\end{aligned}
$}
\end{equation}
which can be further rearranged into:
\begin{equation}
\resizebox{\linewidth}{!}{$
\begin{aligned}
\mathcal{I}_{\epsilon,z}(z')=&-\;\;\;\smashoperator[lr]{\sum_{z_i\in\mathcal{C}_t\cup\mathcal{Z}_{t+1}}}\;\nabla_\theta L(z_i,\hat\theta_{t+1})^\top H_{\hat\theta_{t+1}}^{-1}\nabla_\theta L(z',\hat\theta_{t+1})\\
&+\epsilon\;\smashoperator[lr]{\sum_{z_i\in\mathcal{C}_t\cup\mathcal{Z}_{t+1}}}\;\nabla_\theta L(z_i,\hat\theta_{t+1})^\top H_{\hat\theta_{t+1}}^{-1}H_{\hat\theta_{t+1},z}H_{\hat\theta_{t+1}}^{-1}\nabla_\theta L(z',\hat\theta_{t+1})\\
&-\epsilon\nabla_\theta L(z,\hat\theta_{t+1})^\top H_{\hat\theta_{t+1}}^{-1}\nabla_\theta L(z',\hat\theta_{t+1})\\
&+o(\epsilon).
\end{aligned}
$}
\end{equation}
With notation $s_{t+1}=H_{\hat\theta_{t+1}}^{-1}\sum_{z_i\in\mathcal{C}_t\cup\mathcal{Z}_{t+1}}\nabla_\theta L(z_i,\hat\theta_{t+1})$, its derivative \wrt $\epsilon$ can be written as:
\begin{equation}
\small
\begin{aligned}
&\mathcal{I}^{(2)}(z,z')=\frac{d \mathcal{I}_{\epsilon,z}(z')}{d\epsilon}\Big|_{\epsilon=0}\\
&=-(\nabla_\theta L(z,\hat\theta_{t+1})-H_{\hat\theta_{t+1},z}s_{t+1})^\top H_{\hat\theta_{t+1}}^{-1}\nabla_\theta L(z',\hat\theta_{t+1}).
\end{aligned}
\end{equation}

\section{Intuition behind the deviration}
\begin{figure}[t]
    \centering
    \includegraphics[width=\linewidth]{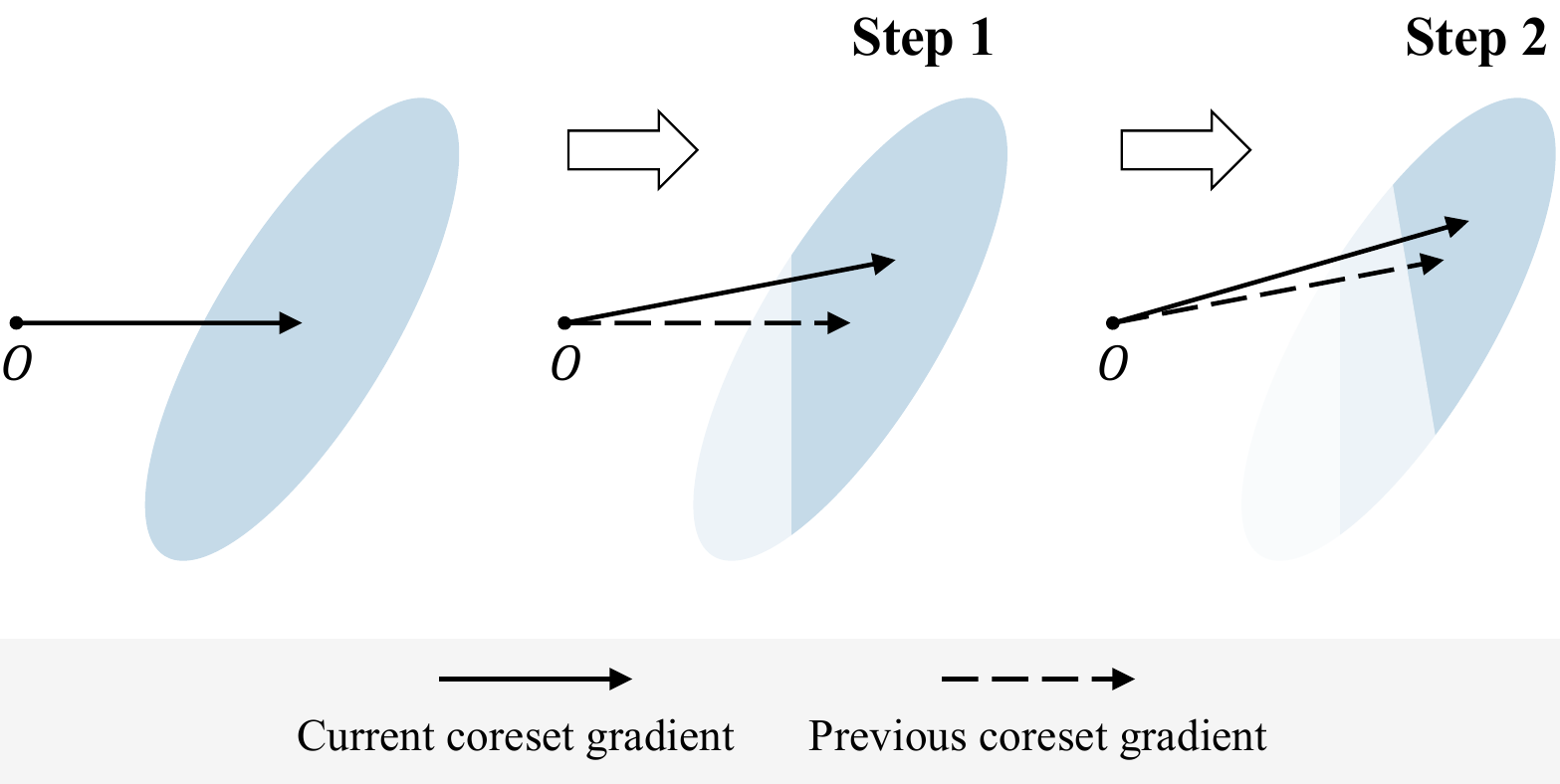}
    \caption{Illustration of two consecutive selection steps based on influence functions. The latter selection turns out to be non-ideal, as evidenced by its decision boundary (between high and low density regions indicated by color intensity) being rotated under the interference of the previous step on gradient information.}
    \label{fig:demo}
\end{figure}
To illustrate the physical meaning behind the equations, this section presents \Cref{fig:demo} as an intuitive example of the second-order effects on sample selection.

It is depicted that after two rounds of selection, the samples are more concentrated in the upper right corner. On a closer look, the prior selection alters the overall gradient, thereby distorting the next selection boundary which is inherently orthogonal to the gradient (by the inner product defined in~\cref{sect:ifs_second}. The final result is thus biased and less diversified.

The illustrated example, which focuses on the drift of decision boundary due to the deviation in coreset gradient, is characterized by our first case of second-order influences in~\cref{eq:if_second_1}. Complementarily, the disturbance to Hessian-related information is tackled in the second case of~\cref{eq:if_second_2}.

\section{Comparison with group influences}
Our second-order influences have a different origin from the group influences proposed by Basu \etal~\cite{basu2020second}. The group effects~\cite{koh2019accuracy, basu2020second} in their work arise from the interaction within a group of reweighted datapoints on the inner objective, so they are limited to jointly optimized samples. Our second-order terms, derived from separate analyses of inner and outer objectives, in contrast, have no such restrictions and apply to sequentially incoming data.

\section{Connection to diversity}
This section presents an algebraic view of the connection between our regularizer and gradient diversity, as a complement to the geometric perspective in~\cref{sect:connection}.

Let $\mathcal{R}^o(\mathcal{C}_t)$ and $\mathcal{R}^i(\mathcal{C}_t)$ denote the regularizers under\linebreak the $\mu=0$ and identical Hessian settings, respectively. They are expressed as:
\begin{equation}
\small
\begin{aligned}
\mathcal{R}^o(\mathcal{C}_t)&=\biggl\|\smashoperator[r]{\sum_{z\in\mathcal{C}_{t-1}\cup\mathcal{Z}_t}}\;\nabla_\theta L(z,\hat\theta_t)-\sum_{z\in\mathcal{C}_t}\nabla_\theta L(z,\hat\theta_t)\biggr\|,\\
\mathcal{R}^i(\mathcal{C}_t)&=\biggl\|(1-\alpha\mu)\;\smashoperator[lr]{\sum_{z\in\mathcal{C}_{t-1}\cup\mathcal{Z}_t}}\;\nabla_\theta L(z,\hat\theta_t)-\sum_{z\in\mathcal{C}_t}\nabla_\theta L(z,\hat\theta_t)\biggr\|.
\end{aligned}
\end{equation}
where $\alpha$ is a coefficient related only to the coreset size. The comparison of the two regularizers yields:
\begin{equation}
\small
\begin{aligned}
\mathcal{R}^i(\mathcal{C}_t)^2&-\mathcal{R}^o(\mathcal{C}_t)^2=\underbrace{(-2\alpha\mu+\alpha^2\mu^2)\biggl\|\smashoperator[r]{\sum_{z\in\mathcal{C}_{t-1}\cup\mathcal{Z}_t}}\;\nabla_\theta L(z,\hat\theta_t)\biggr\|^2}_{\textrm{constant}}\\
&+\underbrace{2\alpha\mu\biggl(\smashoperator[r]{\sum_{z\in\mathcal{C}_{t-1}\cup\mathcal{Z}_t}}\;\nabla_\theta L(z,\hat\theta_t)\biggr)^\top\biggl(\sum_{z\in\mathcal{C}_t}\nabla_\theta L(z,\hat\theta_t)\biggr)}_{\textrm{diversity}},
\end{aligned}
\end{equation}
in which the latter term enforces the coreset gradient to be less aligned with the main gradient. Thus, the regularizer $\mathcal{R}^i(\mathcal{C}_t)$ additionally encourages the inclusion of gradients in other directions and promotes gradient diversity.

\section{Taylor expansion of the regularizer}
To optimize the new equivalent form of our regularizer in~\cref{eq:reg_new}, we perform a first-order Taylor expansion near the initial weight $w_{t,i}^o$:
\begin{equation}
\resizebox{\linewidth}{!}{$
\mathcal{R}(w_t)\approx\mathcal{R}(w_t^o)-\;\smashoperator[lr]{\sum_{z_i\in\mathcal{C}_{t-1}\cup\mathcal{Z}_t}}\;\beta^T(\nabla_\theta L(z_i,\hat\theta_t)-\mu H_{\hat\theta_t,z_i}s_t)(w_{t,i}-w_{t,i}^o),
$}
\end{equation}
where $\beta$ is a vector independent of $w_{t,i}$:
\begin{equation}
\small
\beta=\smashoperator[r]{\sum_{z_i\in\mathcal{C}_{t-1}\cup\mathcal{Z}_t}}\;\frac{(1-w_{t,i}^o)(\nabla_\theta L(z_i,\hat\theta_t)-\mu H_{\hat\theta_t,z_i}s_t)}{\mathcal{R}(w_t^o)}.
\end{equation}
The result is a linear combination of $w_{t,i}$, and thus can be minimized with greedy heuristics, \ie, by iteratively setting the $w_{t,i}$ with the largest coefficient to zero.

\section{Additional results}
\paragraph{Time cost with Hessian-vector product.} The overhead in evaluating the Hessian-vector product is 0.014$\pm$0.001 seconds per step on Split CIFAR-10. This is fairly small compared to the base cost of 0.368$\pm$0.029 seconds per step for computing first-order influence functions.

\begin{table}[t]
    \centering
    \resizebox{\linewidth}{!}{
    \begin{tabular}{lcc}
    \toprule
    Method & Class-incremental & Task-incremental \\
    \midrule
    Grad matching & 39.56$\pm$1.52 $\bullet$ & 88.98$\pm$0.95 $\bullet$\\
    Grad diversity & 43.94$\pm$2.03 $\bullet$ & 87.82$\pm$1.38 $\bullet$\\
    \cmidrule(){1-3}
    Vanilla IF & 47.09$\pm$0.85 $\bullet$ & 90.78$\pm$1.21 $\;\;$\\
    Ours & \textbf{52.81$\pm$1.26} $\;\;$ & \textbf{92.43$\pm$1.11} $\;\;$\\
    \bottomrule
    \end{tabular}
    }
    \caption{Comparison with only gradient regularization, in terms of ACC (\%) on Split CIFAR-10 with $m=500$. $\bullet$ indicates significant improvement with $p$-value less than 0.05 in paired t-tests.}
    \label{tab:reg}
\end{table}
\paragraph{Comparison with only gradient regularization.} Combination of memory replay with gradient regularization based approaches can partly bypass the interference issue. However, it lacks efficiency in buffering the most critical samples for performance preservation. We verify this point through the comparisons in \Cref{tab:reg}, which empirically justifies the motivation of our proposed influence-based scheme.

\begin{table}[t]
    \centering
    \resizebox{\linewidth}{!}{
    \begin{tabular}{lcc}
    \toprule
    Method & Class-incremental & Task-incremental \\
    \midrule
    iCaRL~\cite{rebuffi2017icarl} & 47.87$\pm$0.47 $\bullet$ & 90.35$\pm$1.13 $\;\;$\\
    BiC~\cite{wu2019large} & 51.49$\pm$1.37 $\;\;$ & 90.99$\pm$0.78 $\;\;$\\
    Ours & \textbf{52.81$\pm$1.26} $\;\;$ & \textbf{92.43$\pm$1.11} $\;\;$\\
    \cmidrule(){1-3}\morecmidrules\cmidrule(){1-3}
    ER-ACE~\cite{caccia2022new} & 56.86$\pm$0.64 $\bullet$ & 89.59$\pm$3.23 $\;\;$\\
    ER-ACE + Ours & \textbf{60.57$\pm$0.93} $\;\;$ & \textbf{91.84$\pm$0.71} $\;\;$\\
    \bottomrule
    \end{tabular}
    }
    \caption{Comparison with multi-epoch methods and ER variant in 50-epoch learning. Detailed settings follow \Cref{tab:reg}.}
    \label{tab:cmp}
\end{table}
\paragraph{Comparison with multi-epoch competitors.} Additional comparisons with the classical multi-epoch methods iCaRL~\cite{rebuffi2017icarl} and BiC~\cite{wu2019large} are given in \Cref{tab:cmp}, which confirm the edge of our method in 50-epoch learning. Results are presented with standard deviations.

\paragraph{In combination with ER-ACE.} \Cref{tab:cmp} further tests our strategy on the more advanced replay framework ER-ACE~\cite{caccia2022new} instead of the previously adopted ER~\cite{chaudhry2019tiny}. It is observed that the proposed method combines well with ER-ACE and yields a 3.71\% gain in class-incremental learning.

\paragraph{Additional comparison.} To compare with other replay-based competitors such as OCS~\cite{yoon2022online}, GCR~\cite{yoon2022online} and Bilevel~\cite{borsos2020coresets}, as well as some regularization-based methods such as Stable SGD~\cite{mirzadeh2020understanding} and EWC~\cite{kirkpatrick2017overcoming}, we reimplement our approach using the codebase of OCS. Its framework differs in mainly two aspects: (1) Methods are evaluated on two task-incremental benchmarks, including 20-split CIFAR-100 and a mixture of five datasets from different domains. (2) Each learning stage features much fewer training epochs, so the resulting ACC will be lower than before, while the forgetting metric BWT will be much better.

As shown in~\cref{tab:compare_ocs}, our approach continues to deliver considerable improvement over the base strategy Vanilla IF. Like many replay-based methods, we outperform regularization-based methods by a large margin. Furthermore, our method surpasses the top two competitors OCS and GCR in terms of ACC on both benchmarks. These results again demonstrate the effectiveness of our approach.

\begin{table}[]
    \centering
    \resizebox{\linewidth}{!}{
    \begin{tabular}{lcccc}
    \toprule
    \multirow{2}{*}{Method} & \multicolumn{2}{c}{Split CIFAR-100} & \multicolumn{2}{c}{Multiple Datasets} \\
    \cmidrule(r){2-3} \cmidrule(l){4-5} & ACC (\%) & BWT & ACC (\%) & BWT \\ 
    \midrule
    iCaRL~\cite{rebuffi2017icarl} & 60.3 & -0.04 & - & - \\
    EWC~\cite{kirkpatrick2017overcoming} & 49.5 & -0.48 & 42.7 & -0.28 \\
    A-GEM~\cite{chaudhry2019efficient} & 50.7 & -0.19 & - & - \\
    ER~\cite{chaudhry2019tiny} & 46.9 & -0.21 & - & - \\
    GSS~\cite{aljundi2019gradient} & 59.7 & -0.04 & 60.2 & -0.07 \\
    ER-MIR~\cite{aljundi2019online} & 60.2 & -0.04 & 56.9 & -0.11 \\
    Stable SGD~\cite{mirzadeh2020understanding} & 57.4 & -0.07 & 53.4 & -0.16 \\
    Bilevel~\cite{borsos2020coresets} & 60.1 & -0.04 & 58.1 & -0.08\\
    OCS~\cite{yoon2022online} & 60.5 & -0.04 & 61.5 & \textbf{-0.03} \\
    GCR~\cite{tiwari2022gcr} & 60.9 & - & - & - \\
    \cmidrule{1-5} Vanilla IF & 60.0 & -0.05 & 59.7 & -0.07 \\
    Ours & \textbf{61.2} & \textbf{-0.04} & \textbf{61.6} & -0.05 \\
    \bottomrule
    \end{tabular}
    }
    \caption{Comparison with another group of baseline methods in task-incremental evaluations. The results of most methods come from the summary in OCS~\cite{yoon2022online}, while the result of GCR~\cite{tiwari2022gcr} is provided in its supplementary material.}
    \label{tab:compare_ocs}
\end{table}

\end{document}